\documentclass[sigconf]{acmart}

\usepackage{color}
\usepackage{bm}
\usepackage{cleveref}
\usepackage{booktabs}
\usepackage{setspace}
\usepackage{xspace}
\usepackage{tipa}
\usepackage{color}
\usepackage[ruled]{algorithm2e}
\usepackage{mathtools}  

\usepackage{amsmath,amsfonts,bm}

\def\eqref#1{equation~\ref{#1}}

\def\1{\bm{1}}

\def\vd{{\bm{d}}}

\def\vk{{\bm{k}}}

\def\vp{{\bm{p}}}

\def\mS{{\bm{S}}}

\DeclareMathAlphabet{\mathsfit}{\encodingdefault}{\sfdefault}{m}{sl}
\SetMathAlphabet{\mathsfit}{bold}{\encodingdefault}{\sfdefault}{bx}{n}

\usepackage{amsmath}
\usepackage{amssymb}
\usepackage{mathtools}
\usepackage{amsthm}
\usepackage{microtype}
\usepackage{graphicx}
\usepackage{booktabs} 

\usepackage{subcaption}
\usepackage{multirow}
\usepackage{booktabs}
\usepackage{colortbl}

\AtBeginDocument{%
  \providecommand\BibTeX{{%
    \normalfont B\kern-0.5em{\scshape i\kern-0.25em b}\kern-0.8em\TeX}}}

\copyrightyear{2025}
\acmYear{2025}
\setcopyright{cc}
\setcctype{by}
\acmConference[MM '25]{Proceedings of the 33rd ACM International Conference on Multimedia}{October 27--31, 2025}{Dublin, Ireland}
\acmBooktitle{Proceedings of the 33rd ACM International Conference on Multimedia (MM '25), October 27--31, 2025, Dublin, Ireland}\acmDOI{10.1145/3746027.3755140}
\acmISBN{979-8-4007-2035-2/2025/10}

\definecolor{shadecolor}{rgb}{0.92, 0.92, 0.92}
\definecolor{gtgray}{gray}{0.97}
\definecolor{mygray}{gray}{.88}

\definecolor{gray1}{gray}{.90}
\definecolor{gray2}{gray}{.92}
\definecolor{gray3}{gray}{.94}

\begin{document}

\title{Contextual Gesture: Co-Speech Gesture Video Generation through Context-aware Gesture Representation}

\def\hlinew#1{%
  \noalign{\ifnum0=`}\fi\hrule \@height #1 \futurelet
  \reserved@a\@xhline}

\author{Pinxin Liu}
\email{pliu23@u.rochester}
\orcid{}
\affiliation{%
  \institution{University of Rochester}
  \city{Rochester}
  \state{New York}
  \country{USA}
}

\author{Pengfei Zhang}
\email{pengfz5@uci.edu}
\affiliation{
\institution{University of California Irvine}
  \city{Irvine}
  \state{California}
  \country{USA}
}

\author{Hyeongwoo Kim}
\affiliation{%
  \institution{Imperial College}
  \city{London}
  \state{}
  \country{United Kindom}
}

\author{Pablo Garrido}
\email{pablo.garrido@flawlessai.com}
\affiliation{%
  \institution{Flawless AI}
  \city{Santa Monica}
  \state{California}
  \country{USA}
}

\author{Ari Shapiro}
\email{ariyshapiro@gmail.com}
\affiliation{%
  \institution{Flawless AI}
  \city{Santa Monica}
  \state{California}
  \country{USA}
}

\author{Kyle Olszewski}
\email{olszewski.kyle@gmail.com}
\affiliation{%
  \institution{Flawless AI}
  \city{Santa Monica}
  \state{California}
  \country{USA}
}
\renewcommand{\shortauthors}{Pinxin Liu et al.}

\begin{abstract}

Co-speech gesture generation is crucial for creating lifelike avatars and enhancing human-computer interactions by synchronizing gestures with speech. Despite recent advancements, existing methods struggle with accurately identifying the rhythmic or semantic triggers from audio for generating contextualized gesture patterns and achieving pixel-level realism. To address these challenges, we introduce Contextual Gesture, a framework that improves co-speech gesture video generation through three innovative components: (1) a chronological speech-gesture alignment that temporally connects two modalities, (2) a contextualized gesture tokenization that incorporate speech context into motion pattern representation through distillation, and (3) a structure-aware refinement module that employs edge connection to link gesture keypoints to improve video generation. Our extensive experiments demonstrate that Contextual Gesture not only produces realistic and speech-aligned gesture videos but also supports long-sequence generation and video gesture editing applications, shown in Fig.~\ref{fig:intro}.

\end{abstract}

\vspace{-0.2cm}




\begin{teaserfigure}
\centering
  \includegraphics[width=0.95\textwidth]{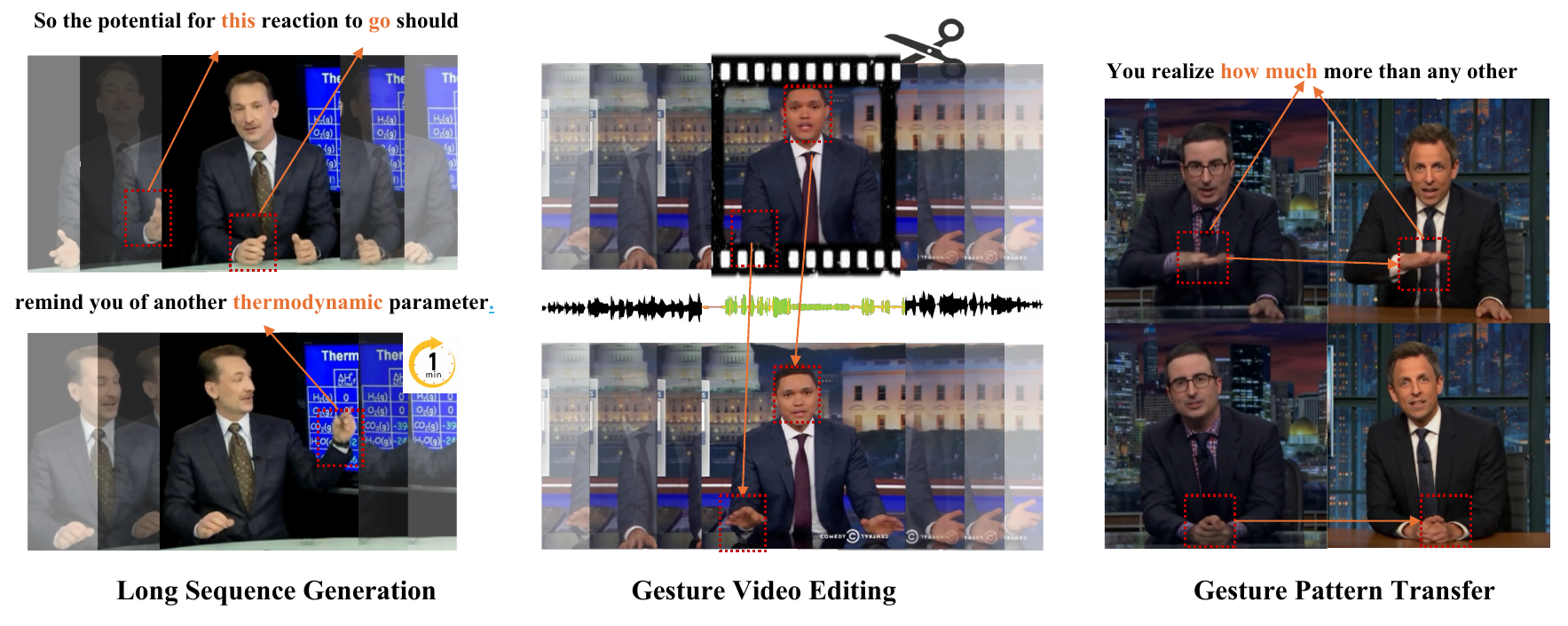}
  \vspace{-0.3cm}
  \caption{Contextual Gesture achieves various fine-grained control over video-level gesture motion. \textbf{Left}: We can generate 30s to 1 min speech conditioned gesture videos. \textbf{Mid}: We modify the gestures for intermediate frames of a video by providing a new audio segment. \textbf{Right}: Different people present the same gesture patterns for a given audio.}
  \label{fig:intro}
\end{teaserfigure}

\maketitle

\section{Introduction}
\label{sec:introduction}

In human communication, speech is often accompanied by gestures that enhance understanding and convey emotions~\cite{de2012interplay}. These non-verbal cues play a crucial role in effective interaction~\cite{burgoon1990nonverbal}, making gesture generation a key component of natural human-computer interaction. Equipping virtual avatars~\cite{song2024texttoon, liu2024gaussianstyle,kinmo} with realistic gesture capabilities is therefore essential for creating engaging interactive experiences for industry production~\cite{tang2025generative}.

Since the era of deep learning paradigm~\cite{zhu2024oftsronestepflowimage,lu2022copy,gao2025eraseanything,Lu_2023_ICCV,Lu_2024_CVPR,ning2024dctdiff}, many recent works tried to model the relationship between the semantic and emotional content of speech, the associated gestures~\cite{yi2022generating, liu2023emage, luo2022much, liu2022learning, liu2022disco, liu2025gesturelsm}. They simplify the problem by generating coarse 3D motion representations—typically keypoints of joints and body parts—that plausibly align with a given speech sample. These simplified representations can be rendered via standard pipelines and effectively capture basic motion patterns. However, they often disregard the speaker’s visual appearance and fine-grained motions, leading to a lack of realism that limits the expressive power and communicative effectiveness of the generated gestures.

Other works, such as ANGIE~\cite{angie} and S2G-diffusion~\cite{s2gdiffusion}, employ image-warping techniques for video generation, guided by motion representations in the form of keypoints obtained from optical-flow-based deformations. While promising, these approaches face several critical limitations.
First, their broad and unconstrained motion representations fail to capture the contextual triggers of gestures—particularly the semantic and emotional cues embedded in speech. This disconnect limits the model’s ability to learn the nuanced relationship between speech and gesture, making it difficult to generate expressive motions that convey the speaker’s intent or metaphoric meaning.
Second, the predicted keypoints primarily encode coarse, large-scale transformations, resulting in unstructured motion patterns that are overly sensitive to pronounced movements. Consequently, the generated outputs often suffer from noise and imprecision, especially in fine-grained areas such as the hands and shoulders.
These issues degrade the realism and coherence of the generated video content.

To address these challenges, we introduce \textit{Contextual Gesture}, a framework designed to generate speech-aligned gesture motions and afterwards high-fidelity speech video. 
To uncover the intrinsic temporal connections between gestures and speech, we employ chronological contrastive learning to align these two modalities into time-sensitive joint representation of speech-contextual and gesture motion, which captures the triggers of gesture patterns influenced by speech. We then use knowledge distillation to incorporate speech-contextual features into the tokenization process of gesture motions, aiming to infuse the implicit intentions of gestures conveyed in the speech. This integration creates a clear chronological linkage between the gestures and the speech, enabling the generation of gestures that reflect the speaker's intentions. 
For motion generation, we employ a masking-based, Transformer gesture generator that produces motion tokens and refines their alignment with the speech signal through bidirectional mask pretraining.
Finally, for uplifting the gesture generation into 2D animations, we propose a structure-aware image refinement module that generates heatmaps of edge connections from keypoints, providing image-level supervision to improve the quality of body regions with large motion.
Extensive experiments demonstrate that our method outperforms the existing approaches in both quantitative and qualitative metrics and achieves long sequence generation and editing applications. In summary, our primary contributions are: 
    
(1)  \textit{Contextual Gesture}, a framework that achieves time-sensitive joint representation of both speech-contextual and gesture motion, and video generation with various application support;

(2) a \textit{Contextual-aware gesture representation} obtained through knowledge distillation from the chronological gesture-speech aligned features from chronological contrastive learning;

(3) a \textit{Structural-aware Image refinement module}, with edge heatmaps as supervision to improve video fidelity.

\section{Related Work}
\label{sec:background}

\begin{figure}
    \includegraphics[width=\linewidth]{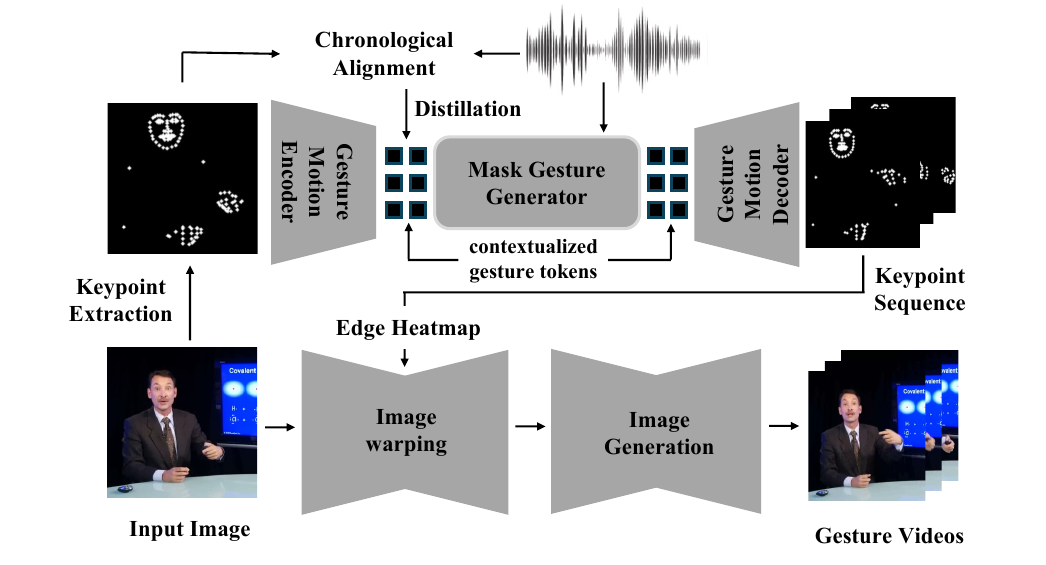}
    \vspace{-2.5em}
    \caption{We extract keypoints and leverage chronological alignment with distillation to achieve contextualized gesture motion representation. We leverage a masking-based generator conditioned on audio to generate gesture keypoint sequences and apply image warping with refinement based on edge heatmaps for final gesture video generation. }
    \label{fig:overview}
    \vspace{-1mm}
\end{figure}

\noindent\textbf{Co-speech Gesture generation}
Co-speech gesture generation is essential for digital avatar production~\cite{song2024tri,kumar2023disjoint} and functions as the prior for realistic rendering~\cite{luo2024splatface,10446837,song2023emotional}. 
\cite{ginosar2019gestures} use an adversarial framework to predict hand and arm poses from audio, and leverage conditional generation~\cite{EverybodyDanceNow} based on pix2pixHD~\cite{wang2018pix2pixHD} for videos. Some recent works~\cite{liu2022learning, Deichler_2023, xu2023chaingenerationmultimodalgesture, ao2022rhythmic,song2021tacr,song2021fsft,liu2025intentionalgesturedeliverintentions,song2024adaptive} learns the hierarchical semantics or leverages contrastive learning to obtain joint audio-gesture embedding from linguistic theory to assist the gesture pose generation. TalkShow~\cite{yi2022generating} estimates SMPL~\cite{SMPL-X:2019} poses, and models the body and hand motions for talk-shows. CaMN~\cite{liu2022beat}, EMAGE~\cite{liu2023emage} and Diffsheg~\cite{diffsheg} use large conversational and speech datasets for joint face and body modeling with diverse style control. ANGIE \cite{angie} and S2G-Diffusion~\cite{s2gdiffusion} use image-warping features based on MRAA \cite{MRAA} or TPS~\cite{zhao2022thin} to model body motion and achieve speech driven animation by learning correspondence between audio and image-warping features. However, none of these works produce structure- and speech-aware motion patterns suitable for achieving natural and realistic gesture rendering.

\vspace{0.2cm}
\noindent\textbf{Conditional Video Generation}
Conditional Video Generation has undergo significant progress with diffusion models~\cite{huang2025fresca,tang2024vidcomposition,dreampose,huang2024scalingconcept,tang2024cardiff}. AnimateDiff~\cite{guo2023animatediff} presents an efficient low-rank adaptation ~\cite{hu2022lora} (LoRA) to adapt image diffusion model for video motion generation. AnimateAnyone~\cite{hu2023animateanyone} construct referencenet for fine-grained control based on skeleton. Make-Your-Anchor~\cite{huang2024makeyouranchor} and Champ~\cite{zhu2024champ} improve avatar video generation through face and body based on SMPL-X conditions. EMO~\cite{tian2024emo} and EchoMimic~\cite{meng2024echomimic} and Tango~\cite{liu2024tango} leverages audio as control signal for talking head and upper body generation. However, these methods are slow in inference speed and ignore the gesture patterns or rhythmic or semantic signals from audio.


\section{Contextual Gesture}

Shown in Fig.~\ref{fig:overview}, our framework targets at generating co-speech photo-realistic human videos with contextualized gestures. To achieve this goal, we first learn time-sensitive contextual-aware gesture representation through knowledge distillation based on chronological gesture-speech alignment (Sec.~\ref{sec:gesture_representation}). We then leverage a Masking-based Gesture Generator for gesture motion generation. (Sec.~\ref{sec:gesture_generation}) To improve the noisy hand and shoulder movement during the transfer of latent motion to pixel space, we propose a structure-aware image refinement through edge heatmaps for guidance. (Sec.~\ref{sec:image-refine}). 

\subsection{Contextualized Gesture Representation}
\label{sec:gesture_representation}

Generating natural and expressive gestures requires capturing fine-grained contextual details that conventional approaches often overlook. Consider a speaker emphasizing the word "really" in the sentence "I really mean it" - while existing methods might generate a generic emphatic gesture, they typically fail to capture the subtle, context-specific body movements that make human gestures truly expressive. This limitation stems from relying solely on motion quantization, which often loses the nuanced relationship between speech and corresponding gestures.

To address this challenge, we propose a novel approach that integrates both audio and semantic information into the motion quantizer's codebook. This integration requires solving two fundamental problems. First, we need to understand how gestures align with speech not just at a high level, but in terms of precise temporal correspondence - when specific words or phrases trigger particular movements, and how the rhythm of speech influences gesture timing. To capture these temporal dynamics, we develop a chronological gesture-speech alignment framework using specialized contrastive learning. Second, we leverage knowledge distillation to incorporate this learned temporal alignment information into the gesture quantizer, enabling our system to generate gestures that are synchronized with speech both semantically and rhythmically.

\begin{figure*}[t!]
	\centering
	\includegraphics[width=1.0\linewidth]{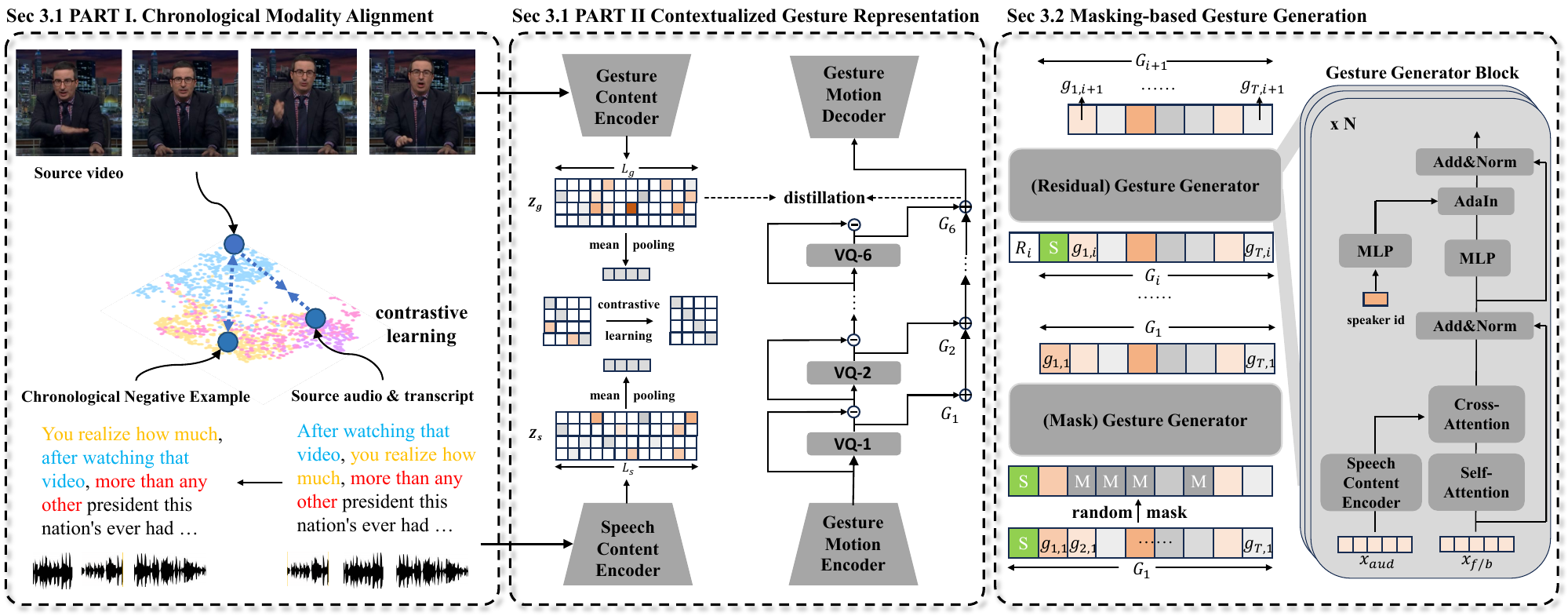}
	\vspace{-0.7cm}
	\caption{For gesture generation, we tokenize gesture keypoint sequences using RVQ-VAE (Sec. 3.1 PART II), and use Gesture Generator to generate the gesture tokens based on masked input (Sec 3.2). The gesture tokens are then used to query the RVQ-VAE in order to generate the gestures. For the RVQ-VAE training, we distill the contextual-aware feature from gesture content encoder (Sec 3.1 PART II), which we trained using chronological contrastive learning (Sec 3.1 PART I), in order to distill speech-motion alignment information into the RVQ-VAE codebook. 
    }
	\label{fig:motion}
	\vspace{-4mm}
\end{figure*}

\vspace{0.2cm}
\noindent \textbf{Feature Representation.} We utilize 2D poses extracted from images to formulate gestures by facial and body movements. We represent a gesture motion sequence as \( G = [F; B] = [f_t; b_t]_{t=1}^{T} \), where \( T \) denotes the length of the motion, \( f \) represents the 2D facial landmarks, and \( b \) denotes the 2D body landmarks. For speech representation, we extract audio embeddings from WavLM~\cite{chen2022wavlm} and Mel spectrogram features~\cite{rabiner2010theory} and beat information using librosa~\cite{mcfee2015librosa}. For text-semantics, we extract embedding from RoBERTa~\cite{roberta}. These features are concatenated to form the speech representation. 

\vspace{0.2cm}
\noindent \textbf{Chronological Speech-Gesture Alignment.}
\label{sec:contrastive}
Traditional approaches to modality alignment~\cite{ao2022rhythmic, liu2022learning, Deichler_2023} rely on global representations through class tokens or max pooling, which overlook the fine-grained temporal dynamics between speech and gestures. We address this limitation by introducing chronological modality alignment.

\textit{\textbf{Vanilla Contrastive Alignment.} }
To align gesture motion patterns with the content of speech and beats, we first project both speech and gesture modalities into a shared embedding space to enhance the speech content awareness of gesture features. As illustrated in Fig.~\ref{fig:motion} Middle, we separately train two gesture content encoders, \(\mathcal{E}_f\) for face motion and \(\mathcal{E}_b\) for body motion, alongside two speech encoders, \(\mathcal{E}_{S_f}\) and \(\mathcal{E}_{S_b}\), to map face and body movements and speech signals into this joint embedding space. For simplicity, we represent the general gesture motion sequence as \( G \). We then apply mean pooling to aggregate content-relevant information to optimize the following loss~\cite{infonce}:
{\small
\begin{equation}
\begin{split} 
    \mathcal{L}_{\text{NCE}} &= - \frac{1}{2N} \sum_{i} \left( \log \frac{\exp{S_{ii}/\tau}}{\sum_{j} \exp{S_{ij}/\tau}} + \log \frac{\exp{S_{ii}/\tau}}{\sum_{j} \exp{S_{ji}/\tau}} \right), ~
\end{split}
\label{eq:infonce}
\end{equation}
}
where \( S \) computes the cosine similarities for all pairs in the batch, defined as \( S_{ij} = \text{cos}(z^s_i, z^g_j) \) and \(\tau\) is the temperature.

\textit{\textbf{Chronological Negative Examples.}} While vanilla Contrastive Learning builds global semantical alignment, we further propose to address the temporal correspondence between speech and gesture. As shown in \cref{fig:motion} Left, consider a speaker saying, "After watching that video, you realize how much, more than any other president...". In this case, the gesture sequence involves "knocking at the table" when saying "more than any other," serving as a visual emphasis for "how much" to highlight the point. To encourage the model understand both semantic and rhythmic alignment between two modalities, we shuffle the words and their corresponding phonemes. By shuffling the sequence to "you realize how much, after watching that video," the semantic intention of the speech is preserved, but the rhythmic correspondence between speech and gesture is disrupted. We use Whisper-X~\cite{bain2022whisperx} to detect temporal segments in the raw sequences. We cut the audio and shuffle these segments, creating these augmented samples as additional chronological negative examples within a batch during contrastive learning.  

\vspace{0.2cm}
\noindent \textbf{Gesture Quantization with Distillation.}
To construct context-aware motion representations, we encode alignment information into the gesture motion codebook. This allows the semantics and contextual triggers from speech to be directly fused into the motion embedding, and enables the generator to easily identify the corresponding motion representation in response to speech triggers. To achieve this goal, we leverage gesture content encoder as the teacher and distill knowledge to codebook latent representation, shown in Fig.~\ref{fig:motion} Middle. We maximize the cosine similarity over time between the RVQ quantization output and the representation from the gesture content encoder:
\begin{equation}
\mathcal{L}_{\text{distill}} = \sum_{t=1}^{T} \cos\left( p(Q_R)^t, Es(G)^t \right)
\end{equation}
%
where \( p \) denotes a linear projection layer, \( Q_R \) is the final quantized output from the RVQ-VAE, \( Es(G) \) represents the output from the gesture content encoder, and \( T \) is the total time frames. The overall training objective is defined as:
\begin{equation}
\mathcal{L}_{\text{rvq}} = \left\lVert x - \hat{x} \right\rVert^2 + \alpha \sum_{r=1}^{R} \left\lVert e_r - \text{sg} \left( z_r - e_r \right) \right\rVert^2 + \beta \mathcal{L}_{\text{distill}}
\end{equation}
where \( \mathcal{L}_{\text{rvq}} \) combines a motion reconstruction loss, a commitment loss~\cite{oord2018neuraldiscreterepresentationlearning} for each layer of quantizer with a distillation loss, with \( \alpha \) and \( \beta \) weighting the contributions.


\subsection{Speech-conditioned Gesture Generation}
\label{sec:gesture_generation}

For Motion Generator, we adopt a similar masking-based generation procedure as in MoMask~\cite{moMask}. We only include the generator data flows in the main paper but defer the training and inference strategy in the Appendix.

\vspace{0.2cm}
\noindent \textbf{Mask Gesture Generator.}
\label{subsec:maskgesture}
As shown in Fig.~\ref{fig:motion} Right, during training, we derive gesture tokens by processing raw gesture sequences through both body and face tokenizers. The gesture token corresponding to the source image acts as the conditioning for all subsequent frames. 
For speech control, we initialize the audio content encoder from alignment pre-training as described in Sec.~\ref{sec:contrastive}. This pre-alignment of gesture tokens with audio encoder features enhances the coherence of gesture generation. We employ cross-attention to integrate audio information and apply Adaptive Instance Normalization (AdaIN)~\cite{adain}, enabling gesture styles based on the speaker's identity.

\vspace{0.2cm}
\noindent \textbf{Residual Gesture Generator.} The Residual Gesture Generator shares a similar architecture with the Masked Gesture Generator, but it includes \( R \) separate embedding layers corresponding to each RVQ residual layer. This module iteratively predicts the residual tokens from the base layers, ultimately producing the final quantized output. Please see MoMask~\cite{moMask} for further details.

\subsection{Structure-Aware Image Generation}
\label{sec:image-refine}
Converting generated gesture motions into realistic videos presents significant challenges, particularly in handling camera movement commonly found in videos. While recent 2D skeleton-based animation methods~\cite{hu2023animateanyone} offer a potential solution, our empirical analysis (detailed in the Appendix) reveals that these approaches struggle with background stability in the presence of camera motion.

To address this limitation, we draw inspiration from optical-flow based image warping methods~\cite{zhao2022thin,MRAA, FOMM}, which have shown promise in handling deformable motion. We adapt these approaches by replacing their unsupervised keypoints with our generated gesture keypoints from Sec.~\ref{sec:gesture_generation} for more precise foreground optical flow estimation. However, this estimation still presents uncertainties, resulting in blurry artifacts around hands and shoulders, particularly when speakers make rapid or large movements.

To address the uncertainties by optical-flow-based deformation during this process, we propose a Structure-Aware Generator. Auto-Link~\cite{autolink} demonstrates that the learning of keypoint connections for image reconstruction aids the model in understanding image semantics. Based on this, we leverage keypoint connections as semantic guidance for image generation.

\vspace{0.2cm}
\noindent \textbf{Edge Heatmaps.}
Using the gesture motion keypoints, we establish linkages between them to provide structural information. To optimize computational efficiency, we limit the number of keypoint connections to those defined by body joint relationships~\cite{wan2017deepskeletonskeletonmap3d}, rather than all potential connections in~\cite{autolink}.

For two keypoints \(\vk_i\) and \(\vk_j\) within connection groups, we create a differentiable edge map \(\mS_{ij}\), modeled as a Gaussian function extending along the line connecting the keypoints. The edge map \(\mS_{ij}\) for keypoints \((\vk_i, \vk_j)\) is defined as:
\begin{equation}
    \mS_{ij}(\vp) = \exp\left(v_{ij}(\vp)d^2_{ij}(\vp) / \sigma^2\right), 
    \label{eq:edge}
\end{equation}
where \(\sigma\) is a learnable parameter controlling the edge thickness, and \(d_{ij}(\vp)\) is the \(L_2\) distance between the pixel \(\vp\) and the edge defined by keypoints \(\vk_i\) and \(\vk_j\):
\begin{equation} 
\vd_{ij}(\vp) = \left\{ \begin{aligned} 
& \lVert \vp-\vk_i\rVert_2 & \text{if}~~t \leq 0, \\ 
& \lVert \vp-((1-t)\vk_i + t\vk_j)\rVert_2 & \text{if}~~0 < t < 1, \\ 
& \lVert \vp-\vk_j\rVert_2 & \text{if}~~t \geq 1, 
\end{aligned}
\right.
\end{equation}

\begin{equation} 
\text{where} \quad t = \frac{(\vp-\vk_i)\cdot (\vk_j-\vk_i)}{\lVert \vk_i-\vk_j\rVert^2_2}.
\end{equation}
Here, \(t\) denotes the normalized distance between \(\vk_i\) and the projection of \(\vp\) onto the edge.

To derive the edge map \(\mS \in \mathbb{R}^{H \times W}\), we take the maximum value at each pixel across all heatmaps:
\begin{equation}
   \mS(\vp)  = \max_{ij} \mS_{ij}(\vp).
   \label{eq:max_heatmap}
\end{equation}


\vspace{0.2cm}
\noindent \textbf{Structural-guided Image Refinement.}  
Traditional optical-flow-based warping methods are effective for handling global deformations but often fail under large motion patterns, such as those involving hands or shoulders, resulting in significant distortions. To address this, we introduce a structure-guided refinement process that incorporates semantic guidance via structural heatmaps.

Instead of directly rendering the warped feature maps into RGB images, we first predict a low-resolution feature map of size \( 256 \times 256 \times 32 \). Multi-resolution edge heatmaps are generated and used as structural cues to refine the feature maps. After performing deformation and occlusion prediction at each scale using TPS~\cite{zhao2022thin}, the edge heatmaps are fed into the generator. Specifically, we integrate these heatmaps into the fusion block using SPADE~\cite{park2019SPADE} and the prediction of residuals are element-wise added to the warped feature maps, ensuring precise structural alignment.

To generate high-resolution RGB images, we employ a U-Net architecture that takes both the warped features and edge heatmaps as inputs. This design preserves fine-grained structural details while compensating for motion-induced distortions. Additional architectural details and analysis are provided in the Appendix.



\vspace{0.2cm}
\noindent \textbf{Training Objective.}
We employ an adversarial loss, along with perceptual similarity loss (LPIPS) ~\cite{lpips} and pixel-level \(L1\) loss for image refinement. The reconstruction objective is defined as:  
\begin{equation}
I_{\text{rec}} = \gamma \mathcal{L}_{GAN} + \mathcal{L}_{L1} + \mathcal{L}_{\text{LPIPS}},
\end{equation}
where \(I_{gt}\) and \(I_{gan}\) represent the ground-truth and generated image separately. We use a small weighted term of $\gamma$ to stablize training.

\begin{figure*}[h]
  \centering
   \includegraphics[width=.9\linewidth]{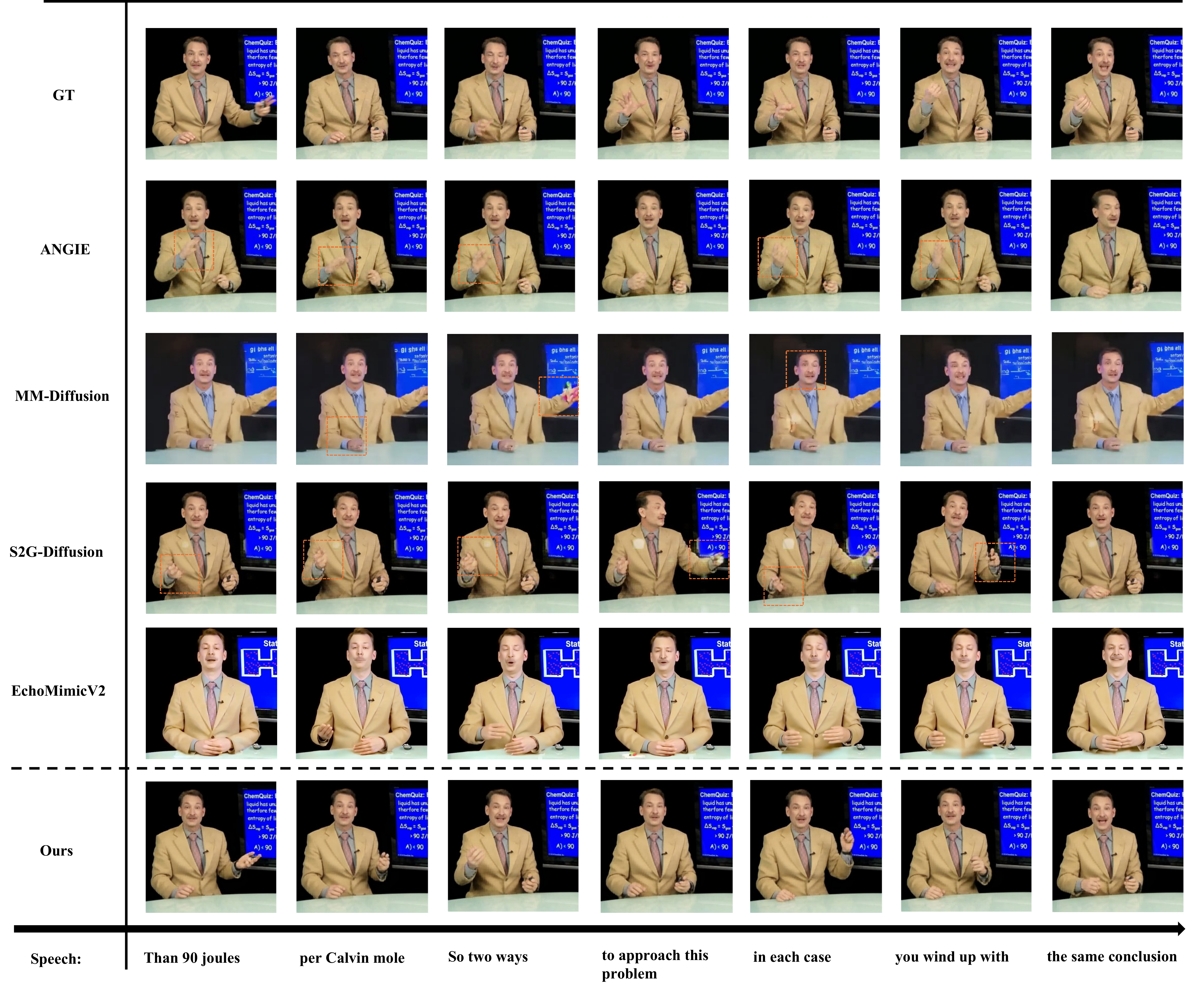}
   \vspace{-0.7cm}
   \caption{\textbf{Visual comparisons.} Our method generates high-quality hand and shoulder motions, and presents metaphoric gestures when saying \textit{``90 joules,''} and \textit{``in each case.''}. Red boxes denote the blurry or unnatural gestures by other methods.}
   \label{fig:compare1}
   \vspace{-0.3cm}
\end{figure*}

\section{Experiments}
\label{sec:exp}
Since our work focuses on joint gesture motion and video generation, to validate the design, we first compare our proposed method for gesture motion generation with relevant 3D gesture motion generation frameworks. We further conduct holistic co-speech gesture video generation comparisons.

\subsection{3D Gesture Motion Generation.}
\noindent\textbf{Dataset.}
We select BEAT-X~\cite{liu2023emage} as the dataset for comparison of gesture generation. For consistency, we exclude the image-to-animation component from our method and leverage 3D SMPL-X poses as in the existing literature. We compare the gesture generation module of our work with representative state-of-the-art methods in co-speech gesture generation~\cite{ao2022rhythmic, yi2022generating, liu2023emage}. We further design a baseline without using contextual distillation.

\vspace{0.2cm}
\noindent\textbf{Evaluation Metrics}
We evaluate the realism of body gestures by Fréchet Gesture Distance (FGD)\cite{yoon2020speech}. We include Diversity by calculating the average L1 distance across clips. For synchronization, we use Beat Constancy (BC)~\cite{li2021ai}. For facial expression, we use the vertex Mean Squared Error (MSE)~\cite{xing2023codetalker} for positional accuracy. 

\vspace{0.2cm}
\noindent\textbf{Experiment Results.}
As shown in \cref{tab:tab4}, our method significantly improves SMPL-X-based co-speech gesture generation, achieving lower FGD, higher diversity. These results indicate that our method produces smoother and more natural gesture motion patterns compared to existing approaches. 
Furthermore, the lower MSE combined with appropriate BC scores indicates that the generated gestures closely track the ground-truth gestures across frames while maintaining rhythmic alignment with the audio. This demonstrates that the gestures are well-coordinated with, and responsive to, the accompanying speech.
These findings demonstrate the effectiveness of our contextual distillation strategy for motion representation learning, as well as the benefit of chronological alignment through contrastive learning. 
We defer the video comparisons in the Appendix videos for reference.

\begin{table}
    \caption{Quantitative results on BEAT-X. 
    FGD (Frechet Gesture Distance) multiplied by \( 10^{-1} \), BC (Beat Constancy) multiplied by \( 10^{-1} \), Diversity, MSE (Mean Squared Error) multiplied by \( 10^{-7} \).
    The best results are in bold. }
    \vspace{-0.3cm}
    \label{tab:tab4}
    \centering
    \setlength{\tabcolsep}{2.5pt}  
    \begin{tabular}{@{}lcccc@{}}  
    \toprule
    Methods & FGD $\downarrow$ & BC $\rightarrow$ & Div. $\uparrow$ & MSE $\downarrow$ \\
    \midrule
    GT &  &  & 0.703 & 11.97 \\
    HA2G~\cite{liu2022learning} & 12.320 & 0.677 & 8.626 & - \\
    DisCo~\cite{liu2022disco} & 9.417 & 0.643 & 9.912 & - \\
    $\text{CaMN}$~\cite{liu2022beat} & 6.644 & 0.676 & 10.86 & - \\
    $\text{DiffSHEG}$~\cite{diffsheg} & 7.141 & 0.743 & 8.21 & 9.571 \\
    $\text{TalkShow}$~\cite{yi2022generating} & 6.209 & 0.695 & 13.47 & 7.791 \\
    $\text{Rhythmic Gesticulator}$~\cite{ao2022rhythmic} & 6.453 & 0.665 & 9.132 & \\
    $\text{EMAGE}$~\cite{liu2023emage} & 5.512 & 0.772 & 13.06 & 7.680 \\
    \midrule
    Ours (w/o distill) & 5.079 & 0.737 & 13.24 & 7.742 \\
    Ours & \textbf{4.434} & 0.724 & \textbf{13.76} & \textbf{7.021} \\
    \bottomrule
    \end{tabular}
    \vspace{-0.55cm}
\end{table}

\subsection{Realistic Video Generation.}
\noindent\textbf{Dataset.}
We utilize PATS~\cite{ginosar2019gestures,ahuja2020no} for the experiments. It contains 84,000 clips from 25 speakers with a mean length of 10.7s, 251 hours in total. For a fair comparison, following the literature~\cite{angie, s2gdiffusion} and replace the missing subject, with 4 speakers are selected (\textit{Noah}, \textit{Kubinec}, \textit{Oliver}, and \textit{Seth}). All video clips are cropped with square bounding boxes, centering speaks, resized to $256\times 256$. We defer the additional details in the Appendix. 

\vspace{0.2cm}
\noindent\textbf{Evaluation Metric.}
For gesture motion metrics, we use \textbf{Fréchet Gesture Distance (FGD)}~\cite{yoon2020speech} to measure the distribution gap between real and generated gestures in feature space, \textbf{Diversity (Div.)}~\cite{lee2019dancing} to calculate the average feature distance between generated gestures, \textbf{Beat Alignment Score (BAS)} following \cite{li2021ai}, \textbf{Percent of
Correct Motion parameters (PCM)}, difference of generation deviate from ground-truth following \cite{diffsheg}. For video generation, we extract 2D human poses for face and body using MMPose~\cite{mmpose2020} to represent the gesture motion. Note that in comparison with other models, the FGD is measued by the keypoints extracted from generated videos in the main experiment while in ablation studies, to prevent the effect image warping errors, FGD is measured by keypoints generated in Sec.\ref{sec:gesture_generation}.

For pixel-level video quality, we assess \textbf{Fréchet Video Distance (FVD)}~\cite{unterthiner2018towards} for the overall quality of gesture videos, \textbf{VQA$_A$} for aesthetics and \textbf{VQA$_T$} for technical quality based on Dover~\cite{wu2023dover}, pretrained on datasets with labels ranked by real users. We further evaluate the training and inference efficiency of various methods, \textbf{Train-T} denotes the number of days for training. \textbf{Infer-T} denotes the number of seconds to produce a 10-second video.

\vspace{0.2cm}
\noindent\textbf{Baseline Methods.}
We benchmark Contextual Gesture against several co-speech gesture video generation methods: (1) ANGIE~\cite{angie}, (2) S2G-Diffusion~\cite{s2gdiffusion}, (3) MM-Diffusion~\cite{ruan2022mmdiffusion}, and
(4) EchoMimicV2~\cite{meng2024echomimic}. The first two are conventional optical-flow-based methods. MM-Diffusion is capable of achieving joint audio-visual generation. EchoMimicV2 is the most recent diffusion based speech-avatar animation model pretrained on large scale data. 

\vspace{0.2cm}
\noindent\textbf{Evaluation Results.}
\label{subsec:quantitative}
We present quantitative evaluations in \cref{tab:comparison}. Our approach significantly outperforms existing methods in both gesture motion and video quality metrics. We provide qualitative evaluations in \cref{fig:compare1}. MM-Diffusion is not able to handle complex motion patterns, leading to almost static results. ANGIE and S2G-Diffusion struggle with local regions, such as the hands, due to its reliance on unsupervised keypoints for global transformations, which neglects local deformations. EchoMimicV2 lacks the background motion modeling and is only capable of presenting the aligned centered avatars in the middle. In addition, it fails to achieve diversified gestures. In contrast, our method demonstrates high-quality video generation, particularly in the facial and body areas. The alignment between gesture and speech is notably enhanced through our speech-content-aware gesture latent representation. For example, when the actor says \textit{”90 joules,”} he points to the screen, and he emphasizes phrases like \textit{”so two ways”} and \textit{”in each case”} by raising his hands.

\begin{table*}
\centering
\caption{Quantitative results shows our method performs better in terms of gesture motions and video generation quality.}
\vspace{-0.4cm}
\label{tab:comparison}
\begin{tabular}{c|cccc|ccc|cc}
\toprule
\multirow{2}{*}{Name}  & \multicolumn{4}{c|}{Gesture Motion Evaluation} & \multicolumn{3}{c|}{Video Quality Assessment} & \multicolumn{2}{c}{Speed}\\ 
& FGD $\downarrow$ & Div. $\uparrow$ & BAS $\uparrow$ & PCM $\uparrow$  & FVD $\downarrow$ & VQA$_A$ $\uparrow$ & VQA$_T$ $\uparrow$ & Train-T $\downarrow$ & Infer-T $\downarrow$\\
\midrule
Ground Truth & 0.0  & 14.01 & 1.00 & 1.00  & 0.00 & 95.69 & 5.33 & - & -\\
\midrule
MM-Diffusion \cite{ruan2022mmdiffusion} & 67.56 & 4.32 & 0.65 & 0.11  & - & 77.65 & 4.14 & 14 days & 600 sec \\
ANGIE \cite{angie} & 34.13 & 7.87 & 0.78 & 0.37  & 515.43 & 86.32 & 4.98 & \textbf{5 days} & 30 sec \\
S2G-Diffusion \cite{s2gdiffusion} & 10.54  & 10.08 & 0.98 & 0.45  & 493.43 & 94.54 & 5.63 & 5 days & 35 sec\\
EchoMimicV2 \cite{meng2024echomimic} & 13.65  & 9.85 & 0.98 & 0.45  & 466.84 & 95.65 & 5.98 & - & 1200 sec\\
\midrule
Ours + AnimateAnyone\cite{hu2023animateanyone} & \textbf{6.56} & 13.06 & 0.99 & 0.54  & 477.82 & 95.63 & 6.04 & 8.5 days & 50 sec\\
Ours & 8.76 & \textbf{13.13} & \textbf{0.99} & \textbf{0.54}  & \textbf{466.43} & \textbf{96.53} & \textbf{6.12} & 6 days & \textbf{3 sec}\\
\bottomrule
\end{tabular}
\vspace{-0.2cm}
\end{table*}

\begin{figure}
\centering
\includegraphics[width=\linewidth]{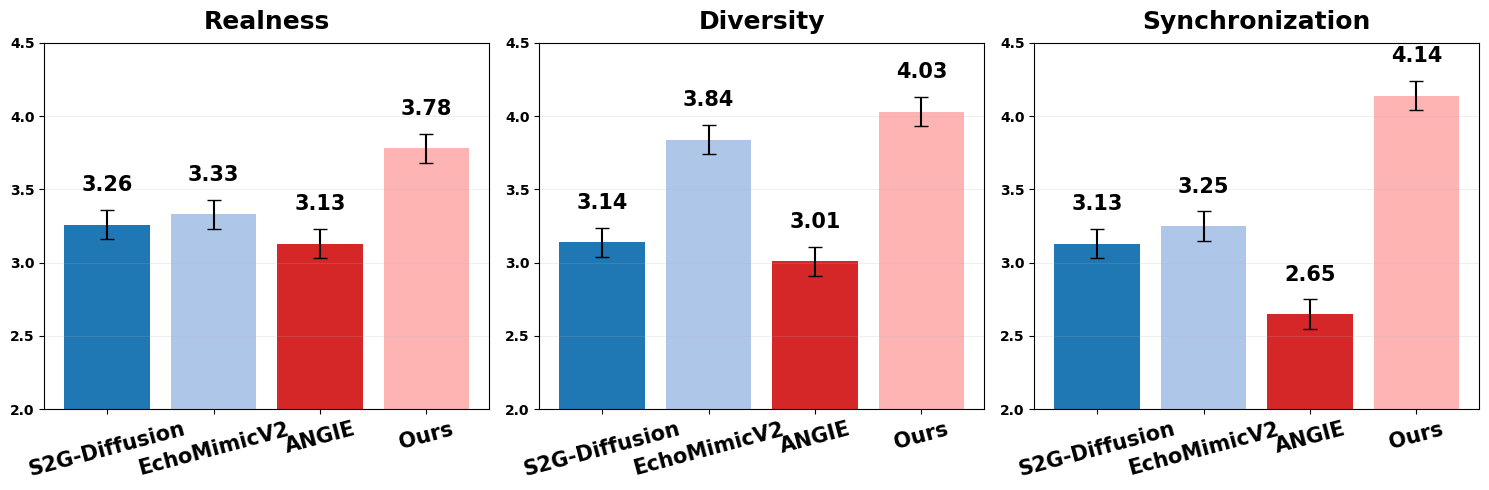}
\vspace{-0.8cm}
\caption{\textbf{User Study}. We generate 80 videos per method for evaluations of \textit{Realness}, \textit{Diversity}, and \textit{Synchronization}.}
 \vspace{-0.75cm}
\label{fig:user-study}
\end{figure}

\vspace{0.2cm}
\noindent\textbf{User Study.}
We conducted a user study to evaluate the visual quality of our method. We sampled 80 videos from each method including EchoMimicV2, S2G-Diffusion, ANGIE and ours and invited 20 participants to conduct Mean Opinion Scores (MOS) evaluations. The rating ranges from 1 (poorest) to 5 (highest). Participants rated the videos on: (1) MOS$_1$: \textit{“How \textbf{realistic} does the video appear?”}, (2) MOS$_2$: \textit{“How \textbf{diverse} does the gesture pattern present?”}, (3) MOS$_3$: \textit{“Are speech and gesture \textbf{synchronized} in this video?”}. The videos were presented in random order to capture participants’ initial impressions. As shown in ~\cref{fig:user-study}, our method outperformed others across realness, synchronization and diversity, achieving significant performance improvement over existing methods. 

\subsection{Ablation Study}
\label{subsec:ablation}
We present ablation studies of keypoint design for image warping, gesture motion representation, generator architecture design, and varios comparisons of image-refinement. We defer additional experiments in the Appendix.

\vspace{0.2cm}
\noindent\textbf{Motion Keypoint Design.}  
We evaluate four settings for image-warping: (1) unsupervised keypoints for global optical-flow transformation (as in ANGIE and S2G-Diffusion), (2) 2D human poses, (3) 2D human poses augmented with flexible learnable points, and (4) full-model reconstruction with refinement. Each design is assessed using TPS~\cite{zhao2022thin} transformation, with self-reconstruction based on these keypoints for evaluation. As shown in \cref{tab:ab_kp}, learnable keypoints lead to a significant decrease in FVD, highlighting their inadequacy for motion control. The inclusion of flexible keypoints does not enhance the image-warping outcomes. Consequently, we opt to utilize 2D pose landmarks exclusively for our study.

\vspace{0.2cm}
\noindent\textbf{Motion Representation.}  
We evaluate several configurations: (1) baseline: no motion representation, relying solely on the generator to synthesize raw 2D landmarks; (2) + RVQ: utilizing Residual VQ (RVQ) to encode joint face-body keypoints; (3) + distill: learning joint embeddings for speech and gesture in both face and body motions; (4) + chrono: leverage chronological alignment for distillation. We discover RVQ significantly improve the precise pose location while distillation leads to natural movements.

\vspace{0.2cm}
\noindent\textbf{Generator Design.}  
We explore various designs for the gesture generator: (1) w/o res: no residual gesture decoder; (2) concat: instead of using cross-attention for audio control, we concatenate the audio features with gesture latent features element-wise during generation; (3) w/o align: the audio encoder is randomly initialized rather than initialized from face and body contrastive learning. Our findings indicate that the Residual Gesture Generator significantly enhances finger motion generation. The cross-attention design outperforms element-wise concatenation, while the pre-alignment of the audio encoder notably improves FGD.

\vspace{0.2cm}
\noindent\textbf{Training Strategies.} We evaluate the mask ratio during training and the number of inference steps during decoding. As shown in \cref{tab:decode}, our model requires only 5 inference steps, in contrast to over 50 or 100 steps in diffusion-based models. Furthermore, a uniform masking ratio between 0.5 and 1 during training yields optimal performance. 

\vspace{0.2cm}
\noindent\textbf{Long Sequence generation.}
To understand the capability of our framework for long sequence generation, we conduct an ablation study for both PATS and BEAT-X dataset. For BEAT-X, we cut the testing audios into segments of 256 (about 8.53 seconds) for short sequence evaluation and use raw testing audios for long sequence evaluation in \cref{tab:tab4}. Shown in \cref{tab:long-seq}, it is interesting for PATS dataset, long-sequence generation as an application in the main paper presents quality lower than normal settings. However, for BEAT-X dataset, the generation quality is not affected much. We attribute this difference caused by the dataset difference. Because PATS dataset consists training video lengths with a average of less than 10 seconds, the model presents less diverse gesture patterns. However, in BEAT-X, most of gesture video sequences are over 30 or 1 minutes, our method further benefits from this long sequence learning precess and presents higher qualities.

\vspace{0.2cm}
\noindent\textbf{Image Refinement.}  
We examine various network designs for motion generation, specifically: (1) w/o refine: no image refinement, relying solely on image warping; (2) + UNet: employing a standard UNet; (3) + pose skeleton: integrating connected skeleton maps as in the diffusion ReferenceNet~\cite{hu2023animateanyone}; (4) + edge heatmap: substituting the previous design with our learnable edge heatmap. Our experiments reveal that the edge heatmap outperforms skeleton maps, due to the learnable thickness of connections for better semantic guidance.

\begin{table*}
  \caption{\textbf{Ablation Studies} for keypoint design, gesture representation, generator architecture, and inference strategies.}
 \vspace{-0.4cm}
  \label{tab:ab}
  \renewcommand{\tabcolsep}{2pt}
  \small
  \begin{subtable}{0.3\linewidth}
    \raggedright
    \begin{tabular}{cccc}
      \toprule
      {\it Kp Repr.} & FVD\(\downarrow\) & LPIPS\(\downarrow\) & PSNR\(\uparrow\)\\
      \midrule
      Unsup-kp & 387.05 & 0.05 & 27.41 \\
      2D-pose & 272.18 & 0.05 & 27.26 \\
      + flex kp & 377.14 & 0.06 & 25.36 \\
      \rowcolor{mygray} full & \textbf{225.77} & \textbf{0.04} & \textbf{27.17}\\
      \bottomrule
    \end{tabular}
    \caption{\small{Configs for keypoint design.}}
    \label{tab:ab_kp}
  \end{subtable}%
  \hfill
  \begin{subtable}{0.22\linewidth}
    \raggedright
    \begin{tabular}{ccc}
      \toprule
      {\it G-Repr.} & FGD\(\downarrow\) & PCM\(\uparrow\)\\
      \midrule
      baseline & 8.84 & 0.35 \\
      +RVQ & 3.43 & 0.37\\
      + distill & 2.75 & 0.58 \\
      \rowcolor{mygray} + chrono & \textbf{0.87} & \textbf{0.59} \\
      \bottomrule
    \end{tabular}
    \caption{\small{Gesture Repr.}}
    \label{tab:ab_motion}
  \end{subtable}%
  \hfill
  \begin{subtable}{0.2\linewidth}
    \raggedright
    \begin{tabular}{ccc}
      \toprule
      {\it G-Gen.} & FGD\(\downarrow\) & PCM\(\uparrow\)\\
      \midrule
      w/o res & 1.62 & 0.51 \\
      concat & 3.55 & 0.51 \\
      w/o align & 1.33 & 0.53 \\
      \rowcolor{mygray} full-model & \textbf{0.87} & \textbf{0.59} \\
      \bottomrule
    \end{tabular}
    \caption{\small{Model Design.}}
    \label{tab:ab_gen}
  \end{subtable}%
  \hfill
  \begin{subtable}{0.24\linewidth}
    \raggedright
    \begin{tabular}{cccc}
      \toprule
      {\it iter.}  & FGD\(\downarrow\)  & Div.\(\uparrow\)  & PCM\(\uparrow\)\\
      \midrule
      \rowcolor{mygray} 5 & \textbf{0.87} & \textbf{13.23} & \textbf{0.59} \\
      10 & 0.98 & 13.11 & 0.57\\
      15 & 1.24 & 13.04 & 0.57\\
      20 & 1.56 & \textbf{13.23} & 0.57\\
      \bottomrule
    \end{tabular}
    \caption{\small{Mask decoding steps.}}
    \label{tab:decode}
  \end{subtable}
  
  \begin{subtable}{0.32\linewidth}
    \raggedright
    \begin{tabular}{cccc}
      \toprule
      {\it M-Ratio} & FGD\(\downarrow\) & Div.\(\uparrow\) & PCM\(\uparrow\)\\
      \midrule
      Uni 0-1 & 2.13 & \textbf{14.31} & 0.56 \\
      Uni .3-1 & 1.56 & 12.44 & 0.512 \\
      \rowcolor{mygray} Uni .5-1 & \textbf{0.87} & 13.23 & \textbf{0.59} \\
      Uni .7-1  & 1.22 & 13.12 & 0.57\\
      \bottomrule
    \end{tabular}
    \caption{\small{Mask-ratio during training.}}
    \label{tab:ab_mask}
  \end{subtable}%
  \hfill
  \begin{subtable}[t]{0.32\linewidth}
    \raggedright
    \begin{tabular}{ccccc}
      \toprule
      Dataset & Setting & FGD\(\downarrow\) & Div.\(\uparrow\) & BAS \\
      \midrule
      \multirow{2}{*}{PATS} & $\leq$10s & 1.303 & 13.260 & 0.996\\
       & $>$10s & 2.356 & 11.956 & 0.994  \\
      \multirow{2}{*}{BEAT-X} & $\leq$10s  & 4.747 & 13.14 & 7.323 \\
       & $>$10s & \textbf{4.650}  & \textbf{13.55} & 7.370\\
      \bottomrule
    \end{tabular}
    \caption{\small{Long Seq Generation Quality.}}
    \label{tab:long-seq}
  \end{subtable}%
  \hfill
  \begin{subtable}[t]{0.32\linewidth}
    \raggedright
    \begin{tabular}{cccc}
      \toprule
      {\it Refine} & VQA$_A$\(\uparrow\) & VQA$_T$\(\uparrow\) & FVD\(\downarrow\)\\
      \midrule
      w/o refine & 92.15 & 5.43 & 494.35 \\
      + UNet & 93.86 & 5.51 & 478.54 \\
      + skeleton & 95.79 & 5.65 & 473.34 \\
      \rowcolor{mygray} + heatmap & \textbf{96.53} & \textbf{6.12} & \textbf{466.43} \\
      \bottomrule
    \end{tabular}
    \caption{\small{Image-refinement strategies.}}
    \label{tab:ab_refine}
  \end{subtable}
  \vspace{-0.3cm}
\end{table*}

\begin{figure*}
  \centering
   \includegraphics[width=.98\linewidth]{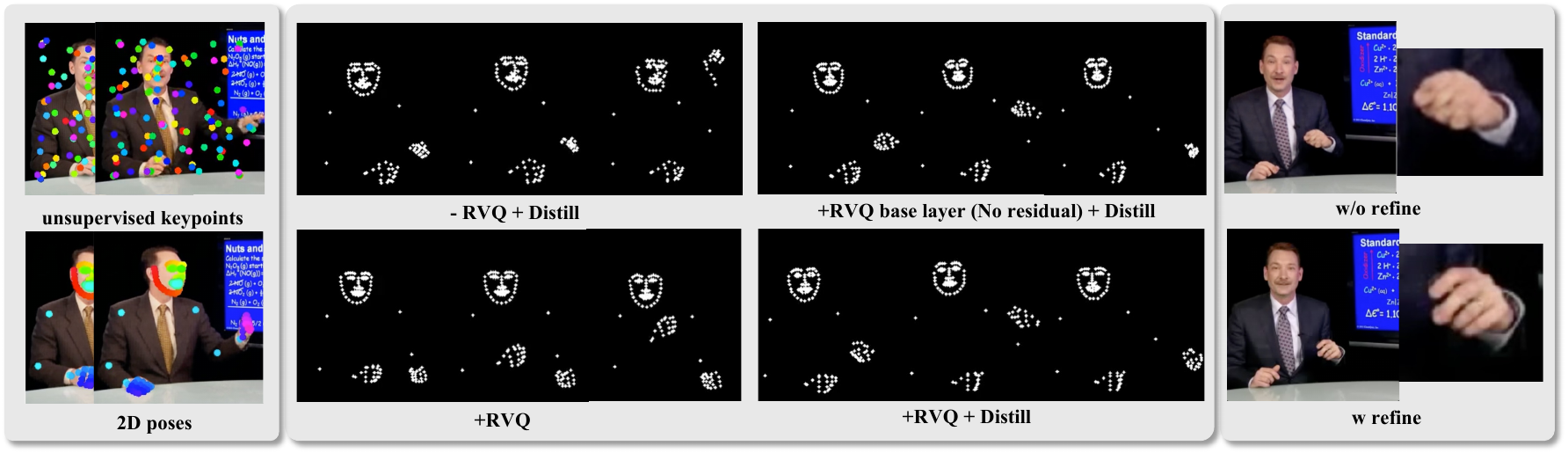}
    \vspace{-0.5cm}
   \caption{\textbf{Ablations.} Left: motion by unsupervised keypoints or 2d poses; 
   Middle: RVQ-based gesture representation and generation; Right: image-refinement helps hand generation.}
   \label{fig:compare2}
   \vspace{-0.1cm}
\end{figure*}

\begin{figure*}
  \centering
   \includegraphics[width=.98\linewidth]{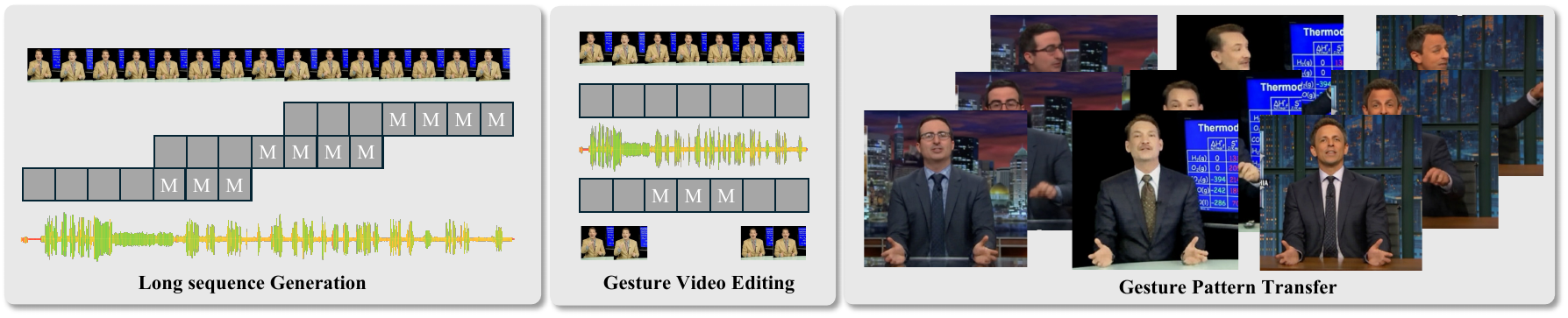}
   \vspace{-0.5cm}
   \caption{\small{Our model supports multiple video gesture generation end editing applications.}}
   \label{fig:application}
   \vspace{-0.3cm}
\end{figure*}

\subsection{Application}

\noindent\textbf{Long Sequence Generation.} As in \cref{fig:application}, to generate long sequences, we begin with the initial frame and corresponding target audio, segmented into smaller windows. After generating the first segment, the last few frames of the output serve as the new starting condition for the next segment, enabling iterative outpainting.

\noindent\textbf{Video Gesture Editing.} For editing, we extract keypoints from the video, tokenize face and body movements into motion tokens, and insert mask tokens where edits are needed. By changing the speech audio or speaker identity, we can create new gesture patterns and re-render the video.

\noindent\textbf{Gesture Pattern Transfer.} With different identity embeddings, we generate unique gesture patterns for the same audio input. See the demo videos in the Appendix.

\noindent\textbf{Speech-Gesture Retrieval.} With chronological speech-gesture alignment, the model is capable of retrieving the best gesture motion correponding to the given speech audio in a batch of data. See additional details in the Appendix.

\vspace{-0.2cm}
\section{Conclusion}
\label{sec:conclusion}

We present \textbf{Contextual Gesture}, a framework for generating realistic co-speech gesture videos.
To ensure the gestures cohere well with speech, we propose speech-content aware gesture motion representation though knowledge distillation from the gesture-speech aligned features.
Our structural-aware image generation module improves the transformation of latent motions into realistic animations for large-scale body motions.
We hope this work encourage further exploration of the relationship between gesture patterns and speech context for better video generations in the future.


\newpage
\bibliographystyle{ACM-Reference-Format}
\bibliography{sample-base}


\begin{thebibliography}{90}


\ifx \showCODEN    \undefined \def \showCODEN     #1{\unskip}     \fi
\ifx \showDOI      \undefined \def \showDOI       #1{#1}\fi
\ifx \showISBNx    \undefined \def \showISBNx     #1{\unskip}     \fi
\ifx \showISBNxiii \undefined \def \showISBNxiii  #1{\unskip}     \fi
\ifx \showISSN     \undefined \def \showISSN      #1{\unskip}     \fi
\ifx \showLCCN     \undefined \def \showLCCN      #1{\unskip}     \fi
\ifx \shownote     \undefined \def \shownote      #1{#1}          \fi
\ifx \showarticletitle \undefined \def \showarticletitle #1{#1}   \fi
\ifx \showURL      \undefined \def \showURL       {\relax}        \fi
\providecommand\bibfield[2]{#2}
\providecommand\bibinfo[2]{#2}
\providecommand\natexlab[1]{#1}
\providecommand\showeprint[2][]{arXiv:#2}

\bibitem[Ahuja et~al\mbox{.}(2020)]%
        {ahuja2020no}
\bibfield{author}{\bibinfo{person}{Chaitanya Ahuja}, \bibinfo{person}{Dong~Won Lee}, \bibinfo{person}{Ryo Ishii}, {and} \bibinfo{person}{Louis-Philippe Morency}.} \bibinfo{year}{2020}\natexlab{}.
\newblock \showarticletitle{{No Gestures Left Behind: Learning Relationships between Spoken Language and Freeform Gestures}}. In \bibinfo{booktitle}{\emph{Proceedings of the 2020 Conference on Empirical Methods in Natural Language Processing: Findings}}. \bibinfo{pages}{1884--1895}.
\newblock


\bibitem[Ao et~al\mbox{.}(2022)]%
        {ao2022rhythmic}
\bibfield{author}{\bibinfo{person}{Tenglong Ao}, \bibinfo{person}{Qingzhe Gao}, \bibinfo{person}{Yuke Lou}, \bibinfo{person}{Baoquan Chen}, {and} \bibinfo{person}{Libin Liu}.} \bibinfo{year}{2022}\natexlab{}.
\newblock \showarticletitle{Rhythmic gesticulator: Rhythm-aware co-speech gesture synthesis with hierarchical neural embeddings}.
\newblock \bibinfo{journal}{\emph{ACM Transactions on Graphics (TOG)}} \bibinfo{volume}{41}, \bibinfo{number}{6} (\bibinfo{year}{2022}), \bibinfo{pages}{1--19}.
\newblock


\bibitem[Bain et~al\mbox{.}(2023)]%
        {bain2022whisperx}
\bibfield{author}{\bibinfo{person}{Max Bain}, \bibinfo{person}{Jaesung Huh}, \bibinfo{person}{Tengda Han}, {and} \bibinfo{person}{Andrew Zisserman}.} \bibinfo{year}{2023}\natexlab{}.
\newblock \showarticletitle{WhisperX: Time-Accurate Speech Transcription of Long-Form Audio}.
\newblock \bibinfo{journal}{\emph{INTERSPEECH 2023}} (\bibinfo{year}{2023}).
\newblock


\bibitem[Burgoon et~al\mbox{.}(1990)]%
        {burgoon1990nonverbal}
\bibfield{author}{\bibinfo{person}{Judee~K Burgoon}, \bibinfo{person}{Thomas Birk}, {and} \bibinfo{person}{Michael Pfau}.} \bibinfo{year}{1990}\natexlab{}.
\newblock \showarticletitle{{Nonverbal Behaviors, Persuasion, and Credibility}}.
\newblock \bibinfo{journal}{\emph{Human communication research}} \bibinfo{volume}{17}, \bibinfo{number}{1} (\bibinfo{year}{1990}), \bibinfo{pages}{140--169}.
\newblock


\bibitem[Chan et~al\mbox{.}(2019)]%
        {EverybodyDanceNow}
\bibfield{author}{\bibinfo{person}{Caroline Chan}, \bibinfo{person}{Shiry Ginosar}, \bibinfo{person}{Tinghui Zhou}, {and} \bibinfo{person}{Alexei~A Efros}.} \bibinfo{year}{2019}\natexlab{}.
\newblock \showarticletitle{{Everybody Dance Now}}. In \bibinfo{booktitle}{\emph{ICCV}}.
\newblock


\bibitem[Chang et~al\mbox{.}(2023)]%
        {chang2023muse}
\bibfield{author}{\bibinfo{person}{Huiwen Chang}, \bibinfo{person}{Han Zhang}, \bibinfo{person}{Jarred Barber}, \bibinfo{person}{AJ Maschinot}, \bibinfo{person}{Jose Lezama}, \bibinfo{person}{Lu Jiang}, \bibinfo{person}{Ming-Hsuan Yang}, \bibinfo{person}{Kevin Murphy}, \bibinfo{person}{William~T Freeman}, \bibinfo{person}{Michael Rubinstein}, {et~al\mbox{.}}} \bibinfo{year}{2023}\natexlab{}.
\newblock \showarticletitle{{Muse: Text-to-Image Generation Via Masked Generative Transformers}}.
\newblock \bibinfo{journal}{\emph{arXiv preprint arXiv:2301.00704}} (\bibinfo{year}{2023}).
\newblock


\bibitem[Chen et~al\mbox{.}(2024)]%
        {diffsheg}
\bibfield{author}{\bibinfo{person}{Junming Chen}, \bibinfo{person}{Yunfei Liu}, \bibinfo{person}{Jianan Wang}, \bibinfo{person}{Ailing Zeng}, \bibinfo{person}{Yu Li}, {and} \bibinfo{person}{Qifeng Chen}.} \bibinfo{year}{2024}\natexlab{}.
\newblock \bibinfo{title}{{DiffSHEG: A Diffusion-Based Approach for Real-Time Speech-driven Holistic 3D Expression and Gesture Generation}}.
\newblock
\newblock
\showeprint[arxiv]{2401.04747}~[cs.SD]
\urldef\tempurl%
\url{https://arxiv.org/abs/2401.04747}
\showURL{%
\tempurl}


\bibitem[Chen et~al\mbox{.}(2022)]%
        {chen2022wavlm}
\bibfield{author}{\bibinfo{person}{Sanyuan Chen}, \bibinfo{person}{Chengyi Wang}, \bibinfo{person}{Zhengyang Chen}, \bibinfo{person}{Yu Wu}, \bibinfo{person}{Shujie Liu}, \bibinfo{person}{Zhuo Chen}, \bibinfo{person}{Jinyu Li}, \bibinfo{person}{Naoyuki Kanda}, \bibinfo{person}{Takuya Yoshioka}, \bibinfo{person}{Xiong Xiao}, {et~al\mbox{.}}} \bibinfo{year}{2022}\natexlab{}.
\newblock \showarticletitle{{WavLM: Large-Scale Self-Supervised Pre-Training for Full Stack Speech Processing}}.
\newblock \bibinfo{journal}{\emph{IEEE Journal of Selected Topics in Signal Processing}} \bibinfo{volume}{16}, \bibinfo{number}{6} (\bibinfo{year}{2022}), \bibinfo{pages}{1505--1518}.
\newblock


\bibitem[De~Ruiter et~al\mbox{.}(2012)]%
        {de2012interplay}
\bibfield{author}{\bibinfo{person}{Jan~P De~Ruiter}, \bibinfo{person}{Adrian Bangerter}, {and} \bibinfo{person}{Paula Dings}.} \bibinfo{year}{2012}\natexlab{}.
\newblock \showarticletitle{{The Interplay Between Gesture and Speech in the Production of Referring Expressions: Investigating the Tradeoff Hypothesis}}.
\newblock \bibinfo{journal}{\emph{Topics in cognitive science}} \bibinfo{volume}{4}, \bibinfo{number}{2} (\bibinfo{year}{2012}), \bibinfo{pages}{232--248}.
\newblock


\bibitem[Deichler et~al\mbox{.}(2023)]%
        {Deichler_2023}
\bibfield{author}{\bibinfo{person}{Anna Deichler}, \bibinfo{person}{Shivam Mehta}, \bibinfo{person}{Simon Alexanderson}, {and} \bibinfo{person}{Jonas Beskow}.} \bibinfo{year}{2023}\natexlab{}.
\newblock \showarticletitle{Diffusion-Based Co-Speech Gesture Generation Using Joint Text and Audio Representation}. In \bibinfo{booktitle}{\emph{INTERNATIONAL CONFERENCE ON MULTIMODAL INTERACTION}} \emph{(\bibinfo{series}{ICMI ’23})}. \bibinfo{publisher}{ACM}.
\newblock
\urldef\tempurl%
\url{https://doi.org/10.1145/3577190.3616117}
\showDOI{\tempurl}


\bibitem[Devlin(2018)]%
        {devlin2018bert}
\bibfield{author}{\bibinfo{person}{Jacob Devlin}.} \bibinfo{year}{2018}\natexlab{}.
\newblock \showarticletitle{{BERT: Pre-Training of Deep Bidirectional Transformers for Language Understanding}}.
\newblock \bibinfo{journal}{\emph{arXiv preprint arXiv:1810.04805}} (\bibinfo{year}{2018}).
\newblock


\bibitem[Gao et~al\mbox{.}(2025)]%
        {gao2025eraseanything}
\bibfield{author}{\bibinfo{person}{Daiheng Gao}, \bibinfo{person}{Shilin Lu}, \bibinfo{person}{Wenbo Zhou}, \bibinfo{person}{Jiaming Chu}, \bibinfo{person}{Jie Zhang}, \bibinfo{person}{Mengxi Jia}, \bibinfo{person}{Bang Zhang}, \bibinfo{person}{Zhaoxin Fan}, {and} \bibinfo{person}{Weiming Zhang}.} \bibinfo{year}{2025}\natexlab{}.
\newblock \showarticletitle{EraseAnything: Enabling Concept Erasure in Rectified Flow Transformers}. In \bibinfo{booktitle}{\emph{Forty-second International Conference on Machine Learning}}.
\newblock
\urldef\tempurl%
\url{https://openreview.net/forum?id=vvBAZJh2nQ}
\showURL{%
\tempurl}


\bibitem[Ginosar et~al\mbox{.}(2019)]%
        {ginosar2019gestures}
\bibfield{author}{\bibinfo{person}{S. Ginosar}, \bibinfo{person}{A. Bar}, \bibinfo{person}{G. Kohavi}, \bibinfo{person}{C. Chan}, \bibinfo{person}{A. Owens}, {and} \bibinfo{person}{J. Malik}.} \bibinfo{year}{2019}\natexlab{}.
\newblock \showarticletitle{{Learning Individual Styles of Conversational Gesture}}. In \bibinfo{booktitle}{\emph{CVPR}}. \bibinfo{publisher}{IEEE}.
\newblock


\bibitem[Guo et~al\mbox{.}(2024a)]%
        {moMask}
\bibfield{author}{\bibinfo{person}{Chuan Guo}, \bibinfo{person}{Yuxuan Mu}, \bibinfo{person}{Muhammad~Gohar Javed}, \bibinfo{person}{Sen Wang}, {and} \bibinfo{person}{Li Cheng}.} \bibinfo{year}{2024}\natexlab{a}.
\newblock \showarticletitle{Momask: Generative masked modeling of 3d human motions}. In \bibinfo{booktitle}{\emph{Proceedings of the IEEE/CVF Conference on Computer Vision and Pattern Recognition}}. \bibinfo{pages}{1900--1910}.
\newblock


\bibitem[Guo et~al\mbox{.}(2024b)]%
        {guo2023animatediff}
\bibfield{author}{\bibinfo{person}{Yuwei Guo}, \bibinfo{person}{Ceyuan Yang}, \bibinfo{person}{Anyi Rao}, \bibinfo{person}{Zhengyang Liang}, \bibinfo{person}{Yaohui Wang}, \bibinfo{person}{Yu Qiao}, \bibinfo{person}{Maneesh Agrawala}, \bibinfo{person}{Dahua Lin}, {and} \bibinfo{person}{Bo Dai}.} \bibinfo{year}{2024}\natexlab{b}.
\newblock \showarticletitle{{AnimateDiff: Animate Your Personalized Text-to-Image Diffusion Models without Specific Tuning}}.
\newblock \bibinfo{journal}{\emph{ICLR}} (\bibinfo{year}{2024}).
\newblock


\bibitem[He et~al\mbox{.}(2022)]%
        {he2022masked}
\bibfield{author}{\bibinfo{person}{Kaiming He}, \bibinfo{person}{Xinlei Chen}, \bibinfo{person}{Saining Xie}, \bibinfo{person}{Yanghao Li}, \bibinfo{person}{Piotr Doll{\'a}r}, {and} \bibinfo{person}{Ross Girshick}.} \bibinfo{year}{2022}\natexlab{}.
\newblock \showarticletitle{{Masked Autoencoders Are Scalable Vision Learners}}. In \bibinfo{booktitle}{\emph{CVPR}}. \bibinfo{pages}{16000--16009}.
\newblock


\bibitem[He et~al\mbox{.}(2024)]%
        {s2gdiffusion}
\bibfield{author}{\bibinfo{person}{Xu He}, \bibinfo{person}{Qiaochu Huang}, \bibinfo{person}{Zhensong Zhang}, \bibinfo{person}{Zhiwei Lin}, \bibinfo{person}{Zhiyong Wu}, \bibinfo{person}{Sicheng Yang}, \bibinfo{person}{Minglei Li}, \bibinfo{person}{Zhiyi Chen}, \bibinfo{person}{Songcen Xu}, {and} \bibinfo{person}{Xiaofei Wu}.} \bibinfo{year}{2024}\natexlab{}.
\newblock \showarticletitle{{Co-Speech Gesture Video Generation via Motion-Decoupled Diffusion Model}}. In \bibinfo{booktitle}{\emph{CVPR}}. \bibinfo{pages}{2263--2273}.
\newblock


\bibitem[He et~al\mbox{.}(2023)]%
        {autolink}
\bibfield{author}{\bibinfo{person}{Xingzhe He}, \bibinfo{person}{Bastian Wandt}, {and} \bibinfo{person}{Helge Rhodin}.} \bibinfo{year}{2023}\natexlab{}.
\newblock \bibinfo{title}{{AutoLink: Self-Supervised Learning of Human Skeletons and Object Outlines by Linking Keypoints}}.
\newblock
\newblock
\showeprint[arxiv]{2205.10636}~[cs.CV]
\urldef\tempurl%
\url{https://arxiv.org/abs/2205.10636}
\showURL{%
\tempurl}


\bibitem[Hu et~al\mbox{.}(2022)]%
        {hu2022lora}
\bibfield{author}{\bibinfo{person}{Edward~J Hu}, \bibinfo{person}{Yelong Shen}, \bibinfo{person}{Phillip Wallis}, \bibinfo{person}{Zeyuan Allen-Zhu}, \bibinfo{person}{Yuanzhi Li}, \bibinfo{person}{Shean Wang}, \bibinfo{person}{Lu Wang}, {and} \bibinfo{person}{Weizhu Chen}.} \bibinfo{year}{2022}\natexlab{}.
\newblock \showarticletitle{{Lo{RA}: Low-Rank Adaptation of Large Language Models}}. In \bibinfo{booktitle}{\emph{ICLR}}.
\newblock
\urldef\tempurl%
\url{https://openreview.net/forum?id=nZeVKeeFYf9}
\showURL{%
\tempurl}


\bibitem[Hu et~al\mbox{.}(2023)]%
        {hu2023animateanyone}
\bibfield{author}{\bibinfo{person}{Li Hu}, \bibinfo{person}{Xin Gao}, \bibinfo{person}{Peng Zhang}, \bibinfo{person}{Ke Sun}, \bibinfo{person}{Bang Zhang}, {and} \bibinfo{person}{Liefeng Bo}.} \bibinfo{year}{2023}\natexlab{}.
\newblock \showarticletitle{{Animate Anyone: Consistent and Controllable Image-to-Video Synthesis for Character Animation}}.
\newblock \bibinfo{journal}{\emph{arXiv preprint arXiv:2311.17117}} (\bibinfo{year}{2023}).
\newblock


\bibitem[Huang et~al\mbox{.}(2025)]%
        {huang2025fresca}
\bibfield{author}{\bibinfo{person}{Chao Huang}, \bibinfo{person}{Susan Liang}, \bibinfo{person}{Yunlong Tang}, \bibinfo{person}{Li Ma}, \bibinfo{person}{Yapeng Tian}, {and} \bibinfo{person}{Chenliang Xu}.} \bibinfo{year}{2025}\natexlab{}.
\newblock \showarticletitle{FreSca: Unveiling the Scaling Space in Diffusion Models}.
\newblock \bibinfo{journal}{\emph{arXiv preprint arXiv:2504.02154}} (\bibinfo{year}{2025}).
\newblock


\bibitem[Huang et~al\mbox{.}(2024a)]%
        {huang2024scalingconcept}
\bibfield{author}{\bibinfo{person}{Chao Huang}, \bibinfo{person}{Susan Liang}, \bibinfo{person}{Yunlong Tang}, \bibinfo{person}{Yapeng Tian}, \bibinfo{person}{Anurag Kumar}, {and} \bibinfo{person}{Chenliang Xu}.} \bibinfo{year}{2024}\natexlab{a}.
\newblock \showarticletitle{Scaling Concept with Text-Guided Diffusion Models}.
\newblock \bibinfo{journal}{\emph{arXiv preprint arXiv:2410.24151}} (\bibinfo{year}{2024}).
\newblock


\bibitem[Huang and Belongie(2017)]%
        {adain}
\bibfield{author}{\bibinfo{person}{Xun Huang} {and} \bibinfo{person}{Serge Belongie}.} \bibinfo{year}{2017}\natexlab{}.
\newblock \bibinfo{title}{{Arbitrary Style Transfer in Real-time with Adaptive Instance Normalization}}.
\newblock
\newblock
\showeprint[arxiv]{1703.06868}~[cs.CV]
\urldef\tempurl%
\url{https://arxiv.org/abs/1703.06868}
\showURL{%
\tempurl}


\bibitem[Huang et~al\mbox{.}(2024b)]%
        {huang2024makeyouranchor}
\bibfield{author}{\bibinfo{person}{Ziyao Huang}, \bibinfo{person}{Fan Tang}, \bibinfo{person}{Yong Zhang}, \bibinfo{person}{Xiaodong Cun}, \bibinfo{person}{Juan Cao}, \bibinfo{person}{Jintao Li}, {and} \bibinfo{person}{Tong-Yee Lee}.} \bibinfo{year}{2024}\natexlab{b}.
\newblock \showarticletitle{Make-Your-Anchor: A Diffusion-based 2D Avatar Generation Framework}.
\newblock \bibinfo{journal}{\emph{arXiv preprint arXiv:2403.16510}} (\bibinfo{year}{2024}).
\newblock


\bibitem[Jiang et~al\mbox{.}(2023)]%
        {rtmpose}
\bibfield{author}{\bibinfo{person}{Tao Jiang}, \bibinfo{person}{Peng Lu}, \bibinfo{person}{Li Zhang}, \bibinfo{person}{Ningsheng Ma}, \bibinfo{person}{Rui Han}, \bibinfo{person}{Chengqi Lyu}, \bibinfo{person}{Yining Li}, {and} \bibinfo{person}{Kai Chen}.} \bibinfo{year}{2023}\natexlab{}.
\newblock \bibinfo{title}{RTMPose: Real-Time Multi-Person Pose Estimation based on MMPose}.
\newblock
\newblock
\urldef\tempurl%
\url{https://doi.org/10.48550/ARXIV.2303.07399}
\showDOI{\tempurl}


\bibitem[Johnson et~al\mbox{.}(2016)]%
        {lpips}
\bibfield{author}{\bibinfo{person}{Justin Johnson}, \bibinfo{person}{Alexandre Alahi}, {and} \bibinfo{person}{Li Fei-Fei}.} \bibinfo{year}{2016}\natexlab{}.
\newblock \bibinfo{title}{{Perceptual Losses for Real-Time Style Transfer and Super-Resolution}}.
\newblock
\newblock
\showeprint[arxiv]{1603.08155}~[cs.CV]
\urldef\tempurl%
\url{https://arxiv.org/abs/1603.08155}
\showURL{%
\tempurl}


\bibitem[Karras et~al\mbox{.}(2023)]%
        {dreampose}
\bibfield{author}{\bibinfo{person}{Johanna Karras}, \bibinfo{person}{Aleksander Holynski}, \bibinfo{person}{Ting-Chun Wang}, {and} \bibinfo{person}{Ira Kemelmacher-Shlizerman}.} \bibinfo{year}{2023}\natexlab{}.
\newblock \showarticletitle{{DreamPose: Fashion Image-to-Video Synthesis via Stable Diffusion}}.
\newblock \bibinfo{journal}{\emph{arXiv preprint arXiv:2304.06025}} (\bibinfo{year}{2023}).
\newblock


\bibitem[Kingma(2014)]%
        {kingma2014adam}
\bibfield{author}{\bibinfo{person}{Diederik~P Kingma}.} \bibinfo{year}{2014}\natexlab{}.
\newblock \showarticletitle{{Adam: A Method for Stochastic Optimization}}.
\newblock \bibinfo{journal}{\emph{arXiv preprint arXiv:1412.6980}} (\bibinfo{year}{2014}).
\newblock


\bibitem[Kumar et~al\mbox{.}(2023)]%
        {kumar2023disjoint}
\bibfield{author}{\bibinfo{person}{Raja Kumar}, \bibinfo{person}{Jiahao Luo}, \bibinfo{person}{Alex Pang}, {and} \bibinfo{person}{James Davis}.} \bibinfo{year}{2023}\natexlab{}.
\newblock \showarticletitle{Disjoint pose and shape for 3d face reconstruction}. In \bibinfo{booktitle}{\emph{Proceedings of the IEEE/CVF International Conference on Computer Vision}}. \bibinfo{pages}{3115--3125}.
\newblock


\bibitem[Lee et~al\mbox{.}(2022)]%
        {rvq}
\bibfield{author}{\bibinfo{person}{Doyup Lee}, \bibinfo{person}{Chiheon Kim}, \bibinfo{person}{Saehoon Kim}, \bibinfo{person}{Minsu Cho}, {and} \bibinfo{person}{Wook-Shin Han}.} \bibinfo{year}{2022}\natexlab{}.
\newblock \bibinfo{title}{{Autoregressive Image Generation Using Residual Quantization}}.
\newblock
\newblock
\showeprint[arxiv]{2203.01941}~[cs.CV]
\urldef\tempurl%
\url{https://arxiv.org/abs/2203.01941}
\showURL{%
\tempurl}


\bibitem[Lee et~al\mbox{.}(2019)]%
        {lee2019dancing}
\bibfield{author}{\bibinfo{person}{Hsin-Ying Lee}, \bibinfo{person}{Xiaodong Yang}, \bibinfo{person}{Ming-Yu Liu}, \bibinfo{person}{Ting-Chun Wang}, \bibinfo{person}{Yu-Ding Lu}, \bibinfo{person}{Ming-Hsuan Yang}, {and} \bibinfo{person}{Jan Kautz}.} \bibinfo{year}{2019}\natexlab{}.
\newblock \showarticletitle{{Dancing to Music}}.
\newblock \bibinfo{journal}{\emph{NeurIPS}}  \bibinfo{volume}{32} (\bibinfo{year}{2019}).
\newblock


\bibitem[Li et~al\mbox{.}(2021)]%
        {li2021ai}
\bibfield{author}{\bibinfo{person}{Ruilong Li}, \bibinfo{person}{Shan Yang}, \bibinfo{person}{David~A Ross}, {and} \bibinfo{person}{Angjoo Kanazawa}.} \bibinfo{year}{2021}\natexlab{}.
\newblock \showarticletitle{{AI Choreographer: Music Conditioned 3D Dance Generation with AIST++}}. In \bibinfo{booktitle}{\emph{Proceedings of the IEEE/CVF International Conference on Computer Vision}}. \bibinfo{pages}{13401--13412}.
\newblock


\bibitem[Li et~al\mbox{.}(2023)]%
        {li2023mage}
\bibfield{author}{\bibinfo{person}{Tianhong Li}, \bibinfo{person}{Huiwen Chang}, \bibinfo{person}{Shlok Mishra}, \bibinfo{person}{Han Zhang}, \bibinfo{person}{Dina Katabi}, {and} \bibinfo{person}{Dilip Krishnan}.} \bibinfo{year}{2023}\natexlab{}.
\newblock \showarticletitle{{MAGE: Masked Generative Encoder to Unify Representation Learning and Image Synthesis}}. In \bibinfo{booktitle}{\emph{CVPR}}. \bibinfo{pages}{2142--2152}.
\newblock


\bibitem[Liu et~al\mbox{.}(2022a)]%
        {liu2022disco}
\bibfield{author}{\bibinfo{person}{Haiyang Liu}, \bibinfo{person}{Naoya Iwamoto}, \bibinfo{person}{Zihao Zhu}, \bibinfo{person}{Zhengqing Li}, \bibinfo{person}{You Zhou}, \bibinfo{person}{Elif Bozkurt}, {and} \bibinfo{person}{Bo Zheng}.} \bibinfo{year}{2022}\natexlab{a}.
\newblock \showarticletitle{{DisCo: Disentangled Implicit Content and Rhythm Learning for Diverse Co-Speech Gestures Synthesis}}. In \bibinfo{booktitle}{\emph{Proceedings of the 30th ACM International Conference on Multimedia}}. \bibinfo{pages}{3764--3773}.
\newblock


\bibitem[Liu et~al\mbox{.}(2024b)]%
        {liu2024tango}
\bibfield{author}{\bibinfo{person}{Haiyang Liu}, \bibinfo{person}{Xingchao Yang}, \bibinfo{person}{Tomoya Akiyama}, \bibinfo{person}{Yuantian Huang}, \bibinfo{person}{Qiaoge Li}, \bibinfo{person}{Shigeru Kuriyama}, {and} \bibinfo{person}{Takafumi Taketomi}.} \bibinfo{year}{2024}\natexlab{b}.
\newblock \showarticletitle{TANGO: Co-Speech Gesture Video Reenactment with Hierarchical Audio Motion Embedding and Diffusion Interpolation}.
\newblock \bibinfo{journal}{\emph{arXiv preprint arXiv:2410.04221}} (\bibinfo{year}{2024}).
\newblock


\bibitem[Liu et~al\mbox{.}(2023)]%
        {liu2023emage}
\bibfield{author}{\bibinfo{person}{Haiyang Liu}, \bibinfo{person}{Zihao Zhu}, \bibinfo{person}{Giorgio Becherini}, \bibinfo{person}{Yichen Peng}, \bibinfo{person}{Mingyang Su}, \bibinfo{person}{You Zhou}, \bibinfo{person}{Naoya Iwamoto}, \bibinfo{person}{Bo Zheng}, {and} \bibinfo{person}{Michael~J Black}.} \bibinfo{year}{2023}\natexlab{}.
\newblock \showarticletitle{{EMAGE: Towards Unified Holistic Co-Speech Gesture Generation via Masked Audio Gesture Modeling}}.
\newblock \bibinfo{journal}{\emph{arXiv preprint arXiv:2401.00374}} (\bibinfo{year}{2023}).
\newblock


\bibitem[Liu et~al\mbox{.}(2022d)]%
        {liu2022beat}
\bibfield{author}{\bibinfo{person}{Haiyang Liu}, \bibinfo{person}{Zihao Zhu}, \bibinfo{person}{Naoya Iwamoto}, \bibinfo{person}{Yichen Peng}, \bibinfo{person}{Zhengqing Li}, \bibinfo{person}{You Zhou}, \bibinfo{person}{Elif Bozkurt}, {and} \bibinfo{person}{Bo Zheng}.} \bibinfo{year}{2022}\natexlab{d}.
\newblock \showarticletitle{{BEAT: A Large-Scale Semantic and Emotional Multi-Modal Dataset for Conversational Gestures Synthesis}}.
\newblock \bibinfo{journal}{\emph{arXiv preprint arXiv:2203.05297}} (\bibinfo{year}{2022}).
\newblock


\bibitem[Liu et~al\mbox{.}(2025a)]%
        {liu2025intentionalgesturedeliverintentions}
\bibfield{author}{\bibinfo{person}{Pinxin Liu}, \bibinfo{person}{Haiyang Liu}, \bibinfo{person}{Luchuan Song}, {and} \bibinfo{person}{Chenliang Xu}.} \bibinfo{year}{2025}\natexlab{a}.
\newblock \bibinfo{title}{Intentional Gesture: Deliver Your Intentions with Gestures for Speech}.
\newblock
\newblock
\showeprint[arxiv]{2505.15197}~[cs.CV]
\urldef\tempurl%
\url{https://arxiv.org/abs/2505.15197}
\showURL{%
\tempurl}


\bibitem[Liu et~al\mbox{.}(2025b)]%
        {liu2025gesturelsm}
\bibfield{author}{\bibinfo{person}{Pinxin Liu}, \bibinfo{person}{Luchuan Song}, \bibinfo{person}{Junhua Huang}, {and} \bibinfo{person}{Chenliang Xu}.} \bibinfo{year}{2025}\natexlab{b}.
\newblock \showarticletitle{GestureLSM: Latent Shortcut based Co-Speech Gesture Generation with Spatial-Temporal Modeling}.
\newblock \bibinfo{journal}{\emph{arXiv preprint arXiv:2501.18898}} (\bibinfo{year}{2025}).
\newblock


\bibitem[Liu et~al\mbox{.}(2024a)]%
        {liu2024gaussianstyle}
\bibfield{author}{\bibinfo{person}{Pinxin Liu}, \bibinfo{person}{Luchuan Song}, \bibinfo{person}{Daoan Zhang}, \bibinfo{person}{Hang Hua}, \bibinfo{person}{Yunlong Tang}, \bibinfo{person}{Huaijin Tu}, \bibinfo{person}{Jiebo Luo}, {and} \bibinfo{person}{Chenliang Xu}.} \bibinfo{year}{2024}\natexlab{a}.
\newblock \showarticletitle{GaussianStyle: Gaussian Head Avatar via StyleGAN}.
\newblock \bibinfo{journal}{\emph{arXiv preprint arXiv:2402.00827}} (\bibinfo{year}{2024}).
\newblock


\bibitem[Liu et~al\mbox{.}(2022b)]%
        {angie}
\bibfield{author}{\bibinfo{person}{Xian Liu}, \bibinfo{person}{Qianyi Wu}, \bibinfo{person}{Hang Zhou}, \bibinfo{person}{Yuanqi Du}, \bibinfo{person}{Wayne Wu}, \bibinfo{person}{Dahua Lin}, {and} \bibinfo{person}{Ziwei Liu}.} \bibinfo{year}{2022}\natexlab{b}.
\newblock \showarticletitle{{Audio-Driven Co-Speech Gesture Video Generation}}.
\newblock \bibinfo{journal}{\emph{NeurIPS}}  \bibinfo{volume}{35} (\bibinfo{year}{2022}), \bibinfo{pages}{21386--21399}.
\newblock


\bibitem[Liu et~al\mbox{.}(2022c)]%
        {liu2022learning}
\bibfield{author}{\bibinfo{person}{Xian Liu}, \bibinfo{person}{Qianyi Wu}, \bibinfo{person}{Hang Zhou}, \bibinfo{person}{Yinghao Xu}, \bibinfo{person}{Rui Qian}, \bibinfo{person}{Xinyi Lin}, \bibinfo{person}{Xiaowei Zhou}, \bibinfo{person}{Wayne Wu}, \bibinfo{person}{Bo Dai}, {and} \bibinfo{person}{Bolei Zhou}.} \bibinfo{year}{2022}\natexlab{c}.
\newblock \showarticletitle{{Learning Hierarchical Cross-Modal Association for Co-Speech Gesture Generation}}. In \bibinfo{booktitle}{\emph{CVPR}}. \bibinfo{pages}{10462--10472}.
\newblock


\bibitem[Liu et~al\mbox{.}(2019)]%
        {roberta}
\bibfield{author}{\bibinfo{person}{Yinhan Liu}, \bibinfo{person}{Myle Ott}, \bibinfo{person}{Naman Goyal}, \bibinfo{person}{Jingfei Du}, \bibinfo{person}{Mandar Joshi}, \bibinfo{person}{Danqi Chen}, \bibinfo{person}{Omer Levy}, \bibinfo{person}{Mike Lewis}, \bibinfo{person}{Luke Zettlemoyer}, {and} \bibinfo{person}{Veselin Stoyanov}.} \bibinfo{year}{2019}\natexlab{}.
\newblock \bibinfo{title}{RoBERTa: A Robustly Optimized BERT Pretraining Approach}.
\newblock
\newblock
\showeprint[arxiv]{1907.11692}~[cs.CL]
\urldef\tempurl%
\url{https://arxiv.org/abs/1907.11692}
\showURL{%
\tempurl}


\bibitem[Lu et~al\mbox{.}(2022)]%
        {lu2022copy}
\bibfield{author}{\bibinfo{person}{Shilin Lu}, \bibinfo{person}{Xinghong Hu}, \bibinfo{person}{Chengyou Wang}, \bibinfo{person}{Lu Chen}, \bibinfo{person}{Shulu Han}, {and} \bibinfo{person}{Yuejia Han}.} \bibinfo{year}{2022}\natexlab{}.
\newblock \showarticletitle{Copy-move image forgery detection based on evolving circular domains coverage}.
\newblock \bibinfo{journal}{\emph{Multimedia Tools and Applications}} \bibinfo{volume}{81}, \bibinfo{number}{26} (\bibinfo{year}{2022}), \bibinfo{pages}{37847--37872}.
\newblock


\bibitem[Lu et~al\mbox{.}(2023)]%
        {Lu_2023_ICCV}
\bibfield{author}{\bibinfo{person}{Shilin Lu}, \bibinfo{person}{Yanzhu Liu}, {and} \bibinfo{person}{Adams Wai-Kin Kong}.} \bibinfo{year}{2023}\natexlab{}.
\newblock \showarticletitle{TF-ICON: Diffusion-Based Training-Free Cross-Domain Image Composition}. In \bibinfo{booktitle}{\emph{Proceedings of the IEEE/CVF International Conference on Computer Vision (ICCV)}}. \bibinfo{pages}{2294--2305}.
\newblock


\bibitem[Lu et~al\mbox{.}(2024)]%
        {Lu_2024_CVPR}
\bibfield{author}{\bibinfo{person}{Shilin Lu}, \bibinfo{person}{Zilan Wang}, \bibinfo{person}{Leyang Li}, \bibinfo{person}{Yanzhu Liu}, {and} \bibinfo{person}{Adams Wai-Kin Kong}.} \bibinfo{year}{2024}\natexlab{}.
\newblock \showarticletitle{MACE: Mass Concept Erasure in Diffusion Models}. In \bibinfo{booktitle}{\emph{Proceedings of the IEEE/CVF Conference on Computer Vision and Pattern Recognition (CVPR)}}. \bibinfo{pages}{6430--6440}.
\newblock


\bibitem[Luo et~al\mbox{.}(2022)]%
        {luo2022much}
\bibfield{author}{\bibinfo{person}{Jiahao Luo}, \bibinfo{person}{Fahim~Hasan Khan}, \bibinfo{person}{Issei Mori}, \bibinfo{person}{Akila de Silva}, \bibinfo{person}{Eric~Sandoval Ruezga}, \bibinfo{person}{Minghao Liu}, \bibinfo{person}{Alex Pang}, {and} \bibinfo{person}{James Davis}.} \bibinfo{year}{2022}\natexlab{}.
\newblock \showarticletitle{How much does input data type impact final face model accuracy?}. In \bibinfo{booktitle}{\emph{Proceedings of the IEEE/CVF Conference on Computer Vision and Pattern Recognition}}. \bibinfo{pages}{18985--18994}.
\newblock


\bibitem[Luo et~al\mbox{.}(2024)]%
        {luo2024splatface}
\bibfield{author}{\bibinfo{person}{Jiahao Luo}, \bibinfo{person}{Jing Liu}, {and} \bibinfo{person}{James Davis}.} \bibinfo{year}{2024}\natexlab{}.
\newblock \showarticletitle{SplatFace: Gaussian splat face reconstruction leveraging an optimizable surface}.
\newblock \bibinfo{journal}{\emph{arXiv preprint arXiv:2403.18784}} (\bibinfo{year}{2024}).
\newblock


\bibitem[Mao et~al\mbox{.}(2024)]%
        {Mao_2024}
\bibfield{author}{\bibinfo{person}{Xiaofeng Mao}, \bibinfo{person}{Zhengkai Jiang}, \bibinfo{person}{Qilin Wang}, \bibinfo{person}{Chencan Fu}, \bibinfo{person}{Jiangning Zhang}, \bibinfo{person}{Jiafu Wu}, \bibinfo{person}{Yabiao Wang}, \bibinfo{person}{Chengjie Wang}, \bibinfo{person}{Wei Li}, {and} \bibinfo{person}{Mingmin Chi}.} \bibinfo{year}{2024}\natexlab{}.
\newblock \showarticletitle{MDT-A2G: Exploring Masked Diffusion Transformers for Co-Speech Gesture Generation}. In \bibinfo{booktitle}{\emph{Proceedings of the 32nd ACM International Conference on Multimedia}} \emph{(\bibinfo{series}{MM ’24})}. \bibinfo{publisher}{ACM}, \bibinfo{pages}{3266–3274}.
\newblock
\urldef\tempurl%
\url{https://doi.org/10.1145/3664647.3680684}
\showDOI{\tempurl}


\bibitem[McFee et~al\mbox{.}(2015)]%
        {mcfee2015librosa}
\bibfield{author}{\bibinfo{person}{Brian McFee}, \bibinfo{person}{Colin Raffel}, \bibinfo{person}{Dawen Liang}, \bibinfo{person}{Daniel~PW Ellis}, \bibinfo{person}{Matt McVicar}, \bibinfo{person}{Eric Battenberg}, {and} \bibinfo{person}{Oriol Nieto}.} \bibinfo{year}{2015}\natexlab{}.
\newblock \showarticletitle{{librosa: Audio and Music Signal Analysis in Python}}. In \bibinfo{booktitle}{\emph{Proceedings of the 14th Python in Science Conference}}, Vol.~\bibinfo{volume}{8}.
\newblock


\bibitem[Meng et~al\mbox{.}(2024)]%
        {meng2024echomimic}
\bibfield{author}{\bibinfo{person}{Rang Meng}, \bibinfo{person}{Xingyu Zhang}, \bibinfo{person}{Yuming Li}, {and} \bibinfo{person}{Chenguang Ma}.} \bibinfo{year}{2024}\natexlab{}.
\newblock \bibinfo{title}{EchoMimicV2: Towards Striking, Simplified, and Semi-Body Human Animation}.
\newblock
\newblock
\showeprint[arxiv]{2411.10061}~[cs.CV]


\bibitem[Ning et~al\mbox{.}(2024)]%
        {ning2024dctdiff}
\bibfield{author}{\bibinfo{person}{Mang Ning}, \bibinfo{person}{Mingxiao Li}, \bibinfo{person}{Jianlin Su}, \bibinfo{person}{Haozhe Jia}, \bibinfo{person}{Lanmiao Liu}, \bibinfo{person}{Martin Bene{\v{s}}}, \bibinfo{person}{Wenshuo Chen}, \bibinfo{person}{Albert~Ali Salah}, {and} \bibinfo{person}{Itir~Onal Ertugrul}.} \bibinfo{year}{2024}\natexlab{}.
\newblock \showarticletitle{Dctdiff: Intriguing properties of image generative modeling in the dct space}.
\newblock \bibinfo{journal}{\emph{arXiv preprint arXiv:2412.15032}} (\bibinfo{year}{2024}).
\newblock


\bibitem[OpenMMLab(2020)]%
        {mmpose2020}
\bibfield{author}{\bibinfo{person}{OpenMMLab}.} \bibinfo{year}{2020}\natexlab{}.
\newblock \bibinfo{title}{{OpenMMLab Pose Estimation Toolbox and Benchmark}}.
\newblock \bibinfo{howpublished}{\url{https://github.com/open-mmlab/mmpose}}.
\newblock


\bibitem[Park et~al\mbox{.}(2019)]%
        {park2019SPADE}
\bibfield{author}{\bibinfo{person}{Taesung Park}, \bibinfo{person}{Ming-Yu Liu}, \bibinfo{person}{Ting-Chun Wang}, {and} \bibinfo{person}{Jun-Yan Zhu}.} \bibinfo{year}{2019}\natexlab{}.
\newblock \showarticletitle{{Semantic Image Synthesis with Spatially-Adaptive Normalization}}. In \bibinfo{booktitle}{\emph{CVPR}}.
\newblock


\bibitem[Pavlakos et~al\mbox{.}(2019)]%
        {SMPL-X:2019}
\bibfield{author}{\bibinfo{person}{Georgios Pavlakos}, \bibinfo{person}{Vasileios Choutas}, \bibinfo{person}{Nima Ghorbani}, \bibinfo{person}{Timo Bolkart}, \bibinfo{person}{Ahmed A.~A. Osman}, \bibinfo{person}{Dimitrios Tzionas}, {and} \bibinfo{person}{Michael~J. Black}.} \bibinfo{year}{2019}\natexlab{}.
\newblock \showarticletitle{{Expressive Body Capture: 3D Hands, Face, and Body from a Single Image}}. In \bibinfo{booktitle}{\emph{CVPR}}.
\newblock


\bibitem[Petrovich et~al\mbox{.}(2023)]%
        {tmr}
\bibfield{author}{\bibinfo{person}{Mathis Petrovich}, \bibinfo{person}{Michael~J. Black}, {and} \bibinfo{person}{Gül Varol}.} \bibinfo{year}{2023}\natexlab{}.
\newblock \bibinfo{title}{TMR: Text-to-Motion Retrieval Using Contrastive 3D Human Motion Synthesis}.
\newblock
\newblock
\showeprint[arxiv]{2305.00976}~[cs.CV]
\urldef\tempurl%
\url{https://arxiv.org/abs/2305.00976}
\showURL{%
\tempurl}


\bibitem[Pinyoanuntapong et~al\mbox{.}(2024)]%
        {pinyoanuntapong2024mmm}
\bibfield{author}{\bibinfo{person}{Ekkasit Pinyoanuntapong}, \bibinfo{person}{Pu Wang}, \bibinfo{person}{Minwoo Lee}, {and} \bibinfo{person}{Chen Chen}.} \bibinfo{year}{2024}\natexlab{}.
\newblock \showarticletitle{{MMM: Generative Masked Motion Model}}. In \bibinfo{booktitle}{\emph{CVPR}}. \bibinfo{pages}{1546--1555}.
\newblock


\bibitem[Rabiner and Schafer(2010)]%
        {rabiner2010theory}
\bibfield{author}{\bibinfo{person}{Lawrence Rabiner} {and} \bibinfo{person}{Ronald Schafer}.} \bibinfo{year}{2010}\natexlab{}.
\newblock \bibinfo{booktitle}{\emph{{Theory and Applications of Digital Speech Processing}}}.
\newblock \bibinfo{publisher}{Prentice Hall Press}.
\newblock


\bibitem[Ruan et~al\mbox{.}(2023)]%
        {ruan2022mmdiffusion}
\bibfield{author}{\bibinfo{person}{Ludan Ruan}, \bibinfo{person}{Yiyang Ma}, \bibinfo{person}{Huan Yang}, \bibinfo{person}{Huiguo He}, \bibinfo{person}{Bei Liu}, \bibinfo{person}{Jianlong Fu}, \bibinfo{person}{Nicholas~Jing Yuan}, \bibinfo{person}{Qin Jin}, {and} \bibinfo{person}{Baining Guo}.} \bibinfo{year}{2023}\natexlab{}.
\newblock \showarticletitle{{MM-Diffusion: Learning Multi-Modal Diffusion Models for Joint Audio and Video Generation}}. In \bibinfo{booktitle}{\emph{CVPR}}.
\newblock


\bibitem[Selvadurai and Selvadurai(2000)]%
        {selvadurai2000biharmonic}
\bibfield{author}{\bibinfo{person}{APS Selvadurai} {and} \bibinfo{person}{APS Selvadurai}.} \bibinfo{year}{2000}\natexlab{}.
\newblock \showarticletitle{{The Biharmonic Equation}}.
\newblock \bibinfo{journal}{\emph{Partial Differential Equations in Mechanics 2: The Biharmonic Equation, Poisson’s Equation}} (\bibinfo{year}{2000}), \bibinfo{pages}{1--502}.
\newblock


\bibitem[Siarohin et~al\mbox{.}(2019)]%
        {FOMM}
\bibfield{author}{\bibinfo{person}{Aliaksandr Siarohin}, \bibinfo{person}{St{\'e}phane Lathuili{\`e}re}, \bibinfo{person}{Sergey Tulyakov}, \bibinfo{person}{Elisa Ricci}, {and} \bibinfo{person}{Nicu Sebe}.} \bibinfo{year}{2019}\natexlab{}.
\newblock \showarticletitle{{First Order Motion Model for Image Animation}}. In \bibinfo{booktitle}{\emph{NeurIPS}}.
\newblock


\bibitem[Siarohin et~al\mbox{.}(2021)]%
        {MRAA}
\bibfield{author}{\bibinfo{person}{Aliaksandr Siarohin}, \bibinfo{person}{Oliver Woodford}, \bibinfo{person}{Jian Ren}, \bibinfo{person}{Menglei Chai}, {and} \bibinfo{person}{Sergey Tulyakov}.} \bibinfo{year}{2021}\natexlab{}.
\newblock \showarticletitle{{Motion Representations for Articulated Animation}}. In \bibinfo{booktitle}{\emph{CVPR}}.
\newblock


\bibitem[Siyao et~al\mbox{.}(2022)]%
        {siyao2022bailando}
\bibfield{author}{\bibinfo{person}{Li Siyao}, \bibinfo{person}{Weijiang Yu}, \bibinfo{person}{Tianpei Gu}, \bibinfo{person}{Chunze Lin}, \bibinfo{person}{Quan Wang}, \bibinfo{person}{Chen Qian}, \bibinfo{person}{Chen~Change Loy}, {and} \bibinfo{person}{Ziwei Liu}.} \bibinfo{year}{2022}\natexlab{}.
\newblock \showarticletitle{{Bailando: 3D Dance Generation by Actor-Critic GPT with Choreographic Memory}}. In \bibinfo{booktitle}{\emph{CVPR}}. \bibinfo{pages}{11050--11059}.
\newblock


\bibitem[Song et~al\mbox{.}(2024a)]%
        {song2024texttoon}
\bibfield{author}{\bibinfo{person}{Luchuan Song}, \bibinfo{person}{Lele Chen}, \bibinfo{person}{Celong Liu}, \bibinfo{person}{Pinxin Liu}, {and} \bibinfo{person}{Chenliang Xu}.} \bibinfo{year}{2024}\natexlab{a}.
\newblock \showarticletitle{Texttoon: Real-time text toonify head avatar from single video}. In \bibinfo{booktitle}{\emph{SIGGRAPH Asia 2024 Conference Papers}}. \bibinfo{pages}{1--11}.
\newblock


\bibitem[Song et~al\mbox{.}(2021a)]%
        {song2021tacr}
\bibfield{author}{\bibinfo{person}{Luchuan Song}, \bibinfo{person}{Bin Liu}, \bibinfo{person}{Guojun Yin}, \bibinfo{person}{Xiaoyi Dong}, \bibinfo{person}{Yufei Zhang}, {and} \bibinfo{person}{Jia-Xuan Bai}.} \bibinfo{year}{2021}\natexlab{a}.
\newblock \showarticletitle{Tacr-net: editing on deep video and voice portraits}. In \bibinfo{booktitle}{\emph{Proceedings of the 29th ACM International Conference on Multimedia}}. \bibinfo{pages}{478--486}.
\newblock


\bibitem[Song et~al\mbox{.}(2024b)]%
        {song2024tri}
\bibfield{author}{\bibinfo{person}{Luchuan Song}, \bibinfo{person}{Pinxin Liu}, \bibinfo{person}{Lele Chen}, \bibinfo{person}{Guojun Yin}, {and} \bibinfo{person}{Chenliang Xu}.} \bibinfo{year}{2024}\natexlab{b}.
\newblock \showarticletitle{Tri 2-plane: Thinking Head Avatar via Feature Pyramid}. In \bibinfo{booktitle}{\emph{European Conference on Computer Vision}}. Springer, \bibinfo{pages}{1--20}.
\newblock


\bibitem[Song et~al\mbox{.}(2024c)]%
        {10446837}
\bibfield{author}{\bibinfo{person}{Luchuan Song}, \bibinfo{person}{Pinxin Liu}, \bibinfo{person}{Guojun Yin}, {and} \bibinfo{person}{Chenliang Xu}.} \bibinfo{year}{2024}\natexlab{c}.
\newblock \showarticletitle{Adaptive Super Resolution for One-Shot Talking-Head Generation}. In \bibinfo{booktitle}{\emph{ICASSP 2024 - 2024 IEEE International Conference on Acoustics, Speech and Signal Processing (ICASSP)}}. \bibinfo{pages}{4115--4119}.
\newblock
\urldef\tempurl%
\url{https://doi.org/10.1109/ICASSP48485.2024.10446837}
\showDOI{\tempurl}


\bibitem[Song et~al\mbox{.}(2024d)]%
        {song2024adaptive}
\bibfield{author}{\bibinfo{person}{Luchuan Song}, \bibinfo{person}{Pinxin Liu}, \bibinfo{person}{Guojun Yin}, {and} \bibinfo{person}{Chenliang Xu}.} \bibinfo{year}{2024}\natexlab{d}.
\newblock \showarticletitle{Adaptive super resolution for one-shot talking-head generation}. In \bibinfo{booktitle}{\emph{ICASSP 2024-2024 IEEE International Conference on Acoustics, Speech and Signal Processing (ICASSP)}}. IEEE, \bibinfo{pages}{4115--4119}.
\newblock


\bibitem[Song et~al\mbox{.}(2023)]%
        {song2023emotional}
\bibfield{author}{\bibinfo{person}{Luchuan Song}, \bibinfo{person}{Guojun Yin}, \bibinfo{person}{Zhenchao Jin}, \bibinfo{person}{Xiaoyi Dong}, {and} \bibinfo{person}{Chenliang Xu}.} \bibinfo{year}{2023}\natexlab{}.
\newblock \showarticletitle{Emotional listener portrait: Realistic listener motion simulation in conversation}. In \bibinfo{booktitle}{\emph{2023 IEEE/CVF International Conference on Computer Vision (ICCV)}}. IEEE, \bibinfo{pages}{20782--20792}.
\newblock


\bibitem[Song et~al\mbox{.}(2021b)]%
        {song2021fsft}
\bibfield{author}{\bibinfo{person}{Luchuan Song}, \bibinfo{person}{Guojun Yin}, \bibinfo{person}{Bin Liu}, \bibinfo{person}{Yuhui Zhang}, {and} \bibinfo{person}{Nenghai Yu}.} \bibinfo{year}{2021}\natexlab{b}.
\newblock \showarticletitle{Fsft-net: face transfer video generation with few-shot views}. In \bibinfo{booktitle}{\emph{2021 IEEE international conference on image processing (ICIP)}}. IEEE, \bibinfo{pages}{3582--3586}.
\newblock


\bibitem[Tang et~al\mbox{.}(2024a)]%
        {tang2024vidcomposition}
\bibfield{author}{\bibinfo{person}{Yunlong Tang}, \bibinfo{person}{Junjia Guo}, \bibinfo{person}{Hang Hua}, \bibinfo{person}{Susan Liang}, \bibinfo{person}{Mingqian Feng}, \bibinfo{person}{Xinyang Li}, \bibinfo{person}{Rui Mao}, \bibinfo{person}{Chao Huang}, \bibinfo{person}{Jing Bi}, \bibinfo{person}{Zeliang Zhang}, {et~al\mbox{.}}} \bibinfo{year}{2024}\natexlab{a}.
\newblock \showarticletitle{VidComposition: Can MLLMs Analyze Compositions in Compiled Videos?}
\newblock \bibinfo{journal}{\emph{arXiv preprint arXiv:2411.10979}} (\bibinfo{year}{2024}).
\newblock


\bibitem[Tang et~al\mbox{.}(2025)]%
        {tang2025generative}
\bibfield{author}{\bibinfo{person}{Yunlong Tang}, \bibinfo{person}{Junjia Guo}, \bibinfo{person}{Pinxin Liu}, \bibinfo{person}{Zhiyuan Wang}, \bibinfo{person}{Hang Hua}, \bibinfo{person}{Jia-Xing Zhong}, \bibinfo{person}{Yunzhong Xiao}, \bibinfo{person}{Chao Huang}, \bibinfo{person}{Luchuan Song}, \bibinfo{person}{Susan Liang}, {et~al\mbox{.}}} \bibinfo{year}{2025}\natexlab{}.
\newblock \showarticletitle{Generative AI for Cel-Animation: A Survey}.
\newblock \bibinfo{journal}{\emph{arXiv preprint arXiv:2501.06250}} (\bibinfo{year}{2025}).
\newblock


\bibitem[Tang et~al\mbox{.}(2024b)]%
        {tang2024cardiff}
\bibfield{author}{\bibinfo{person}{Yunlong Tang}, \bibinfo{person}{Gen Zhan}, \bibinfo{person}{Li Yang}, \bibinfo{person}{Yiting Liao}, {and} \bibinfo{person}{Chenliang Xu}.} \bibinfo{year}{2024}\natexlab{b}.
\newblock \showarticletitle{Cardiff: Video salient object ranking chain of thought reasoning for saliency prediction with diffusion}.
\newblock \bibinfo{journal}{\emph{arXiv preprint arXiv:2408.12009}} (\bibinfo{year}{2024}).
\newblock


\bibitem[Tevet et~al\mbox{.}(2022)]%
        {mdm}
\bibfield{author}{\bibinfo{person}{Guy Tevet}, \bibinfo{person}{Sigal Raab}, \bibinfo{person}{Brian Gordon}, \bibinfo{person}{Yonatan Shafir}, \bibinfo{person}{Daniel Cohen-Or}, {and} \bibinfo{person}{Amit~H Bermano}.} \bibinfo{year}{2022}\natexlab{}.
\newblock \showarticletitle{{Human Motion Diffusion Model}}.
\newblock \bibinfo{journal}{\emph{arXiv preprint arXiv:2209.14916}} (\bibinfo{year}{2022}).
\newblock


\bibitem[Tian et~al\mbox{.}(2024)]%
        {tian2024emo}
\bibfield{author}{\bibinfo{person}{Linrui Tian}, \bibinfo{person}{Qi Wang}, \bibinfo{person}{Bang Zhang}, {and} \bibinfo{person}{Liefeng Bo}.} \bibinfo{year}{2024}\natexlab{}.
\newblock \showarticletitle{{EMO: Emote Portrait Alive-Generating Expressive Portrait Videos with Audio2Video Diffusion Model Under Weak Conditions}}.
\newblock \bibinfo{journal}{\emph{arXiv preprint arXiv:2402.17485}} (\bibinfo{year}{2024}).
\newblock


\bibitem[Unterthiner et~al\mbox{.}(2018)]%
        {unterthiner2018towards}
\bibfield{author}{\bibinfo{person}{Thomas Unterthiner}, \bibinfo{person}{Sjoerd Van~Steenkiste}, \bibinfo{person}{Karol Kurach}, \bibinfo{person}{Raphael Marinier}, \bibinfo{person}{Marcin Michalski}, {and} \bibinfo{person}{Sylvain Gelly}.} \bibinfo{year}{2018}\natexlab{}.
\newblock \showarticletitle{{Towards Accurate Generative Models of Video: A New Metric \& Challenges}}.
\newblock \bibinfo{journal}{\emph{arXiv preprint arXiv:1812.01717}} (\bibinfo{year}{2018}).
\newblock


\bibitem[van~den Oord et~al\mbox{.}(2019)]%
        {infonce}
\bibfield{author}{\bibinfo{person}{Aaron van~den Oord}, \bibinfo{person}{Yazhe Li}, {and} \bibinfo{person}{Oriol Vinyals}.} \bibinfo{year}{2019}\natexlab{}.
\newblock \bibinfo{title}{Representation Learning with Contrastive Predictive Coding}.
\newblock
\newblock
\showeprint[arxiv]{1807.03748}~[cs.LG]
\urldef\tempurl%
\url{https://arxiv.org/abs/1807.03748}
\showURL{%
\tempurl}


\bibitem[van~den Oord et~al\mbox{.}(2018)]%
        {oord2018neuraldiscreterepresentationlearning}
\bibfield{author}{\bibinfo{person}{Aaron van~den Oord}, \bibinfo{person}{Oriol Vinyals}, {and} \bibinfo{person}{Koray Kavukcuoglu}.} \bibinfo{year}{2018}\natexlab{}.
\newblock \bibinfo{title}{{Neural Discrete Representation Learning}}.
\newblock
\newblock
\showeprint[arxiv]{1711.00937}~[cs.LG]
\urldef\tempurl%
\url{https://arxiv.org/abs/1711.00937}
\showURL{%
\tempurl}


\bibitem[Wan et~al\mbox{.}(2017)]%
        {wan2017deepskeletonskeletonmap3d}
\bibfield{author}{\bibinfo{person}{Qingfu Wan}, \bibinfo{person}{Wei Zhang}, {and} \bibinfo{person}{Xiangyang Xue}.} \bibinfo{year}{2017}\natexlab{}.
\newblock \bibinfo{title}{{DeepSkeleton: Skeleton Map for 3D Human Pose Regression}}.
\newblock
\newblock
\showeprint[arxiv]{1711.10796}~[cs.CV]
\urldef\tempurl%
\url{https://arxiv.org/abs/1711.10796}
\showURL{%
\tempurl}


\bibitem[Wang(2023)]%
        {wang2023t2mhifigptgeneratinghighquality}
\bibfield{author}{\bibinfo{person}{Congyi Wang}.} \bibinfo{year}{2023}\natexlab{}.
\newblock \bibinfo{title}{T2M-HiFiGPT: Generating High Quality Human Motion from Textual Descriptions with Residual Discrete Representations}.
\newblock
\newblock
\showeprint[arxiv]{2312.10628}~[cs.CV]
\urldef\tempurl%
\url{https://arxiv.org/abs/2312.10628}
\showURL{%
\tempurl}


\bibitem[Wang et~al\mbox{.}(2018)]%
        {wang2018pix2pixHD}
\bibfield{author}{\bibinfo{person}{Ting-Chun Wang}, \bibinfo{person}{Ming-Yu Liu}, \bibinfo{person}{Jun-Yan Zhu}, \bibinfo{person}{Andrew Tao}, \bibinfo{person}{Jan Kautz}, {and} \bibinfo{person}{Bryan Catanzaro}.} \bibinfo{year}{2018}\natexlab{}.
\newblock \showarticletitle{{High-Resolution Image Synthesis and Semantic Manipulation with Conditional GANs}}. In \bibinfo{booktitle}{\emph{CVPR}}.
\newblock


\bibitem[Wu et~al\mbox{.}(2023)]%
        {wu2023dover}
\bibfield{author}{\bibinfo{person}{Haoning Wu}, \bibinfo{person}{Erli Zhang}, \bibinfo{person}{Liang Liao}, \bibinfo{person}{Chaofeng Chen}, \bibinfo{person}{Jingwen~Hou Hou}, \bibinfo{person}{Annan Wang}, \bibinfo{person}{Wenxiu~Sun Sun}, \bibinfo{person}{Qiong Yan}, {and} \bibinfo{person}{Weisi Lin}.} \bibinfo{year}{2023}\natexlab{}.
\newblock \showarticletitle{{Exploring Video Quality Assessment on User Generated Contents from Aesthetic and Technical Perspectives}}. In \bibinfo{booktitle}{\emph{ICCV}}.
\newblock


\bibitem[Xing et~al\mbox{.}(2023)]%
        {xing2023codetalker}
\bibfield{author}{\bibinfo{person}{Jinbo Xing}, \bibinfo{person}{Menghan Xia}, \bibinfo{person}{Yuechen Zhang}, \bibinfo{person}{Xiaodong Cun}, \bibinfo{person}{Jue Wang}, {and} \bibinfo{person}{Tien-Tsin Wong}.} \bibinfo{year}{2023}\natexlab{}.
\newblock \showarticletitle{Codetalker: Speech-driven 3d facial animation with discrete motion prior}. In \bibinfo{booktitle}{\emph{Proceedings of the IEEE/CVF Conference on Computer Vision and Pattern Recognition}}. \bibinfo{pages}{12780--12790}.
\newblock


\bibitem[Xu et~al\mbox{.}(2023)]%
        {xu2023chaingenerationmultimodalgesture}
\bibfield{author}{\bibinfo{person}{Zunnan Xu}, \bibinfo{person}{Yachao Zhang}, \bibinfo{person}{Sicheng Yang}, \bibinfo{person}{Ronghui Li}, {and} \bibinfo{person}{Xiu Li}.} \bibinfo{year}{2023}\natexlab{}.
\newblock \bibinfo{title}{Chain of Generation: Multi-Modal Gesture Synthesis via Cascaded Conditional Control}.
\newblock
\newblock
\showeprint[arxiv]{2312.15900}~[cs.CV]
\urldef\tempurl%
\url{https://arxiv.org/abs/2312.15900}
\showURL{%
\tempurl}


\bibitem[Yi et~al\mbox{.}(2023)]%
        {yi2022generating}
\bibfield{author}{\bibinfo{person}{Hongwei Yi}, \bibinfo{person}{Hualin Liang}, \bibinfo{person}{Yifei Liu}, \bibinfo{person}{Qiong Cao}, \bibinfo{person}{Yandong Wen}, \bibinfo{person}{Timo Bolkart}, \bibinfo{person}{Dacheng Tao}, {and} \bibinfo{person}{Michael~J Black}.} \bibinfo{year}{2023}\natexlab{}.
\newblock \showarticletitle{{Generating Holistic 3D Human Motion from Speech}}. In \bibinfo{booktitle}{\emph{CVPR}}.
\newblock


\bibitem[Yoon et~al\mbox{.}(2020)]%
        {yoon2020speech}
\bibfield{author}{\bibinfo{person}{Youngwoo Yoon}, \bibinfo{person}{Bok Cha}, \bibinfo{person}{Joo-Haeng Lee}, \bibinfo{person}{Minsu Jang}, \bibinfo{person}{Jaeyeon Lee}, \bibinfo{person}{Jaehong Kim}, {and} \bibinfo{person}{Geehyuk Lee}.} \bibinfo{year}{2020}\natexlab{}.
\newblock \showarticletitle{{Speech Gesture Generation from the Trimodal Context of Text, Audio, and Speaker Identity}}.
\newblock \bibinfo{journal}{\emph{ACM TOG}} \bibinfo{volume}{39}, \bibinfo{number}{6} (\bibinfo{year}{2020}).
\newblock


\bibitem[Zhang et~al\mbox{.}(2025)]%
        {kinmo}
\bibfield{author}{\bibinfo{person}{Pengfei Zhang}, \bibinfo{person}{Pinxin Liu}, \bibinfo{person}{Hyeongwoo Kim}, \bibinfo{person}{Pablo Garrido}, {and} \bibinfo{person}{Bindita Chaudhuri}.} \bibinfo{year}{2025}\natexlab{}.
\newblock \bibinfo{title}{KinMo: Kinematic-aware Human Motion Understanding and Generation}.
\newblock
\newblock
\showeprint[arxiv]{2411.15472}~[cs.CV]
\urldef\tempurl%
\url{https://arxiv.org/abs/2411.15472}
\showURL{%
\tempurl}


\bibitem[Zhao and Zhang(2022)]%
        {zhao2022thin}
\bibfield{author}{\bibinfo{person}{Jian Zhao} {and} \bibinfo{person}{Hui Zhang}.} \bibinfo{year}{2022}\natexlab{}.
\newblock \showarticletitle{{Thin-Plate Spline Motion Model for Image Animation}}. In \bibinfo{booktitle}{\emph{CVPR}}. \bibinfo{pages}{3657--3666}.
\newblock


\bibitem[Zhu et~al\mbox{.}(2024a)]%
        {zhu2024champ}
\bibfield{author}{\bibinfo{person}{Shenhao Zhu}, \bibinfo{person}{Junming~Leo Chen}, \bibinfo{person}{Zuozhuo Dai}, \bibinfo{person}{Yinghui Xu}, \bibinfo{person}{Xun Cao}, \bibinfo{person}{Yao Yao}, \bibinfo{person}{Hao Zhu}, {and} \bibinfo{person}{Siyu Zhu}.} \bibinfo{year}{2024}\natexlab{a}.
\newblock \showarticletitle{{Champ: Controllable and Consistent Human Image Animation with 3D Parametric Guidance}}.
\newblock \bibinfo{journal}{\emph{arXiv preprint arXiv:2403.14781}} (\bibinfo{year}{2024}).
\newblock


\bibitem[Zhu et~al\mbox{.}(2024b)]%
        {zhu2024oftsronestepflowimage}
\bibfield{author}{\bibinfo{person}{Yuanzhi Zhu}, \bibinfo{person}{Ruiqing Wang}, \bibinfo{person}{Shilin Lu}, \bibinfo{person}{Junnan Li}, \bibinfo{person}{Hanshu Yan}, {and} \bibinfo{person}{Kai Zhang}.} \bibinfo{year}{2024}\natexlab{b}.
\newblock \bibinfo{title}{OFTSR: One-Step Flow for Image Super-Resolution with Tunable Fidelity-Realism Trade-offs}.
\newblock
\newblock
\showeprint[arxiv]{2412.09465}~[cs.CV]
\urldef\tempurl%
\url{https://arxiv.org/abs/2412.09465}
\showURL{%
\tempurl}


\end{thebibliography}

\clearpage
\setcounter{page}{1}
\onecolumn
\appendix

\begin{center} 
    \centering
    \textbf{\large Contextual Gesture: Co-Speech Gesture Video Generation Through Context-Aware Gesture Representation}
\end{center}
\begin{center} 
    \centering
    \large Supplementary Material
\end{center}

\section{Overview}
\label{sec:Summary}
The supplementary material is organized into the following sections:
\begin{itemize}
    \item Section \ref{sec:sub_related}:  Additional Related Works
    \item Section \ref{sec:sub_dataset}:  Dataset Details and Preprocessing
    \item Section \ref{sec:sub-implement}: Additional Implementation Details
    \item Section \ref{sec:s2g-generation}: Gesture Motion Generation
    \item Section \ref{sec:s2g-alignment}: Gesture Speech Retrieval
    \item Section \ref{sec:sub-exp}: Additional Experiments
    \item Section \ref{sec:sub-time}: Time and Resource Consumption
    \item Section \ref{sec:sub_user}: User Study Details
    \item Section \ref{sec:sub-tps} TPS-based Image Warping
    \item Section \ref{sec:ethics} Ethical Ethical Considerations
    \item Section \ref{sec:limitation}: Limitations

\end{itemize}
For more visualization, please see the additional demo videos.

\section{Additional Related Works}
\label{sec:sub_related}
\paragraph{Masked Representation Learning for Generation}
Masked Representation Learning has been demonstated an effective representation learning for various modalities.~\cite{devlin2018bert, he2022masked} Some works explored the generation capabilities using this paradigm. MAGE~\cite{li2023mage} achieves high-quality image generation through iterative remasking. Muse~\cite{chang2023muse} extends this idea to leverage language with region masking for image editing and achieve fine-grained control. Recent Masking Models~\cite{pinyoanuntapong2024mmm, moMask, wang2023t2mhifigptgeneratinghighquality, Mao_2024,kinmo} bring this strategy to the motion and gesture domain and improves the motion generation speed, quality, and editing capability. Inspired by these work, we propose the masked gesture generation conditioned the audio to learn the gesture-speech correspondence during generation.

\section{Dataset Details and Preprocessing}
\label{sec:sub_dataset}

\subsection{Preprocessing}
We found that many videos used in ANGIE~\cite{angie} and S2G-Diffusion~\cite{s2gdiffusion}, particularly for the subject \textit{Jon}, are no longer available. To address this, we replaced \textit{Jon} with \textit{Noah}. We utilized the PATS~\cite{ginosar2019gestures} metadata to download videos from YouTube and preprocess them. After filtering, we obtained 1080 videos for \textit{Oliver}, 1080 for \textit{Kubinec}, 1080 for \textit{Seth}, and 988 for \textit{Noah}. For the testing dataset, we collected 120 videos for \textit{Oliver}, 120 for \textit{Kubinec}, 120 for \textit{Seth}, and 94 for \textit{Noah}.

During the dataset preprocessing, while for image-generation we use the whole video preprocessed as above, for for the speech-gesture alignment and gesture pattern generation modules, we further preprocess the data by slicing them into smaller chunks following S2G-Diffusion~\cite{s2gdiffusion}. Specifically, based on the source training dataset, the keypoint sequences and audio sequences are clipped to 80 frames (3.2s) with stride 10 (0.4s) for training. We obtain 85971 overlapping training examples and 8867 testing examples for gesture pattern modeling.

\subsection{Feature Representation}
\paragraph{Gesture Keypoints.} 
We utilize RTMPose~\cite{rtmpose} from MMPose~\cite{mmpose2020} for whole-human-body keypoint identification. The keypoint definition is based on by 133 CoCo human pose estimation. Due to the PATS~\cite{ginosar2019gestures} only contains the upper body, we select 68 face landmarks for face motion modeling, 3 for left shoulder, 3 for right shoulder, 21 for left hand and 21 for right hand separately, which results in flattened face feature with dim of 136 and body feature with dim of 96.

\paragraph{Audio Features.} 
The audio features are pre-extracted WavLM features (dim of 1024) with additional low-level mel-spectrum and beat information with dimension of 34. We concatenate them channel-wise as the speech feature.

\subsection{Dataset License.}
The video data within PATS dataset include personal identity information, and we strictly adhere to the data usage license ``CC BY - NC - ND 4.0 International,'' which permits non-commercial use.

\section{Additional Implementation Details}
\label{sec:sub-implement}

We jointly train the framework on three speakers. The following sections provide the details for each module's training.

\paragraph{Optimizer Settings.}
All modules utilize the Adam Optimizer~\cite{kingma2014adam} during training, with a learning rate of \(1 \times 10^{-4}\), \(\beta_1 = 0.5\), and \(\beta_2 = 0.999\).

\paragraph{Speech-Gesture Alignment.}
For aligning speech with facial and bodily gestures, we implement two standard transformer blocks for encoding each modality. The latent dimension is configured to 384, accompanied by a feedforward size of 1024. We calculate the mean features for both modalities and project them using a two-layer MLP in a contrastive learning framework, with a temperature parameter set to 0.7.

\paragraph{Residual Vector Quantization (RVQ) Tokenization.}

The overall training objective for the RVQ-VAE is defined as:
\begin{equation}
\mathcal{L}_{\text{rvq}} = \mathbb{E}_{x \sim p(x)} \left[ \left\lVert x - \hat{x} \right\rVert^2 \right] + \alpha \sum_{r=1}^{R} \mathbb{E}_{z_r \sim q(z_r|x)} \left[ \left\lVert e_r - \text{sg} \left( z_r - e_r \right) \right\lVert^2 \right] + \beta \mathcal{L}_{\text{distill}}
\end{equation}
where \( \mathcal{L}_{\text{rvq}} \) combines a motion reconstruction loss, a commitment loss~\cite{oord2018neuraldiscreterepresentationlearning} for each layer of quantizer with a distillation loss, with \( \alpha \) and \( \beta \) weighting the contributions.

We employ six layers of codebooks for residual vector quantization~\cite{rvq} for both face and body modalities, each comprising 1024 codes. To address potential collapse issues during training, we implement codebook resets. The RVQ encoder and decoder are built with two layers of convolutional blocks and a latent dimension of 128. We avoid temporal down-sampling to ensure the latent features maintain the same temporal length as the original input sequences. During RVQ training, we set \(\alpha = 1\) and \(\beta = 0.5\) to balance gesture reconstruction with speech-context distillation.

\paragraph{Mask Gesture Generator.}
The generator takes sequences of discrete tokens for both face and body, derived from the RVQ codebook. This module includes two layers of audio encoders for face and body, initialized based on the Speech-Gesture Alignment. The latent dimension is again set to 384, with a feedforward dimension of 1024, and it features eight layers for both modalities. A two-layer MLP is utilized to project the latent space to the codebook dimension, and cross-entropy is employed for model training. We calculate reconstruction and acceleration loss by feeding the predicted tokens into the RVQ decoder. A reconstruction loss of 50 is maintained during training, and the mask ratio is uniformly varied between 0.5 and 1.0. For inference, a cosine schedule is adopted for decoding. 
We uniformly mask between 50\% and 100\% of the tokens during training. Following the BERT~\cite{devlin2018bert}, when a token is selected for masking, we replace it with a \texttt{[MASK]} token 80\% of the time, a random token 10\% of the time, and leave it unchanged 10\% of the time.  
The Mask Gesture Generator is trained over 250 epochs, taking approximately 1 days to complete.

\paragraph{Residual Gesture Generator.}
The Residual Gesture Generator is designed similarly to the Mask Gesture Generator but utilizes only six layers for the generator. It features four embedding and classification layers corresponding to the RVQ tokenization scheme for residual layers. During training, we randomly select a quantizer layer \( j \in [1, R] \) for learning. All tokens from the preceding layers \( t^{0:j-1} \) are embedded and summed to form the token embedding input. After generating the base layer predictions of discrete tokens from the Masked Gesture Generator, these tokens are fed into the Residual Gesture Generator. This module iteratively predicts the tokens from the base layers, ultimately producing the final quantized output. This module is trained for an additional 500 epochs, requiring about 0.5 days to finalize.


\paragraph{Image Warping.}
For pixel-level motion generation, we utilize Thin Plate Splines (TPS)~\cite{zhao2022thin}. Our framework tracks 116 keypoints (68 for the face and 48 for the body). The number of TPS transformations \(K\) is set to 29, with each transformation utilizing \(N = 4\) paired keypoints. In accordance with TPS methodologies, both the dense motion network and occlusion-aware generators leverage 2D convolutions to produce \(64 \times 64\) weight maps for optical flow generation, along with four occlusion masks at various resolutions (32, 64, 128, and 256) to facilitate image frame synthesis.

\paragraph{Structure-aware Image-Refinement.} We use the UNet similar to S2G-Diffusion~\cite{s2gdiffusion} to restore missing details, further improve the hand and shoulder areas. We keep the training loss to be the same except the added conditional adversarial loss based on edge heatmap. For the network design difference, we add the multi-level edge heatmap as additional control for different resolutions (32, 64 and 128). Each correponds to a SPADE~\cite{park2019SPADE} block to inject the semantic control into the current generation.


\section{Gesture Motion Generation}
\label{sec:s2g-generation}
\paragraph{Inference.} While existing works~\cite{liu2023emage, yi2022generating, diffsheg} leverage auto-regressive next-token prediction or diffusion-based generation process, these strategies hinder the fast synthesis for real-time applications. To resolve this problem, as in Fig.~\ref{fig:motion}, we employ an iterative mask prediction strategy to decode motion tokens during inference. Initially, all tokens are masked except for the first token from the source frame. Conditioned on the audio input, the Mask Gesture Generator predicts probabilities for the masked tokens. In the \( l \)-th iteration, the tokens with the lowest confidence are re-masked, while the remaining tokens stay unchanged for subsequent iterations. This updated sequence continues to inform predictions until the final iteration, when the base-layer tokens are fully generated. Upon completion, the Residual Gesture Generator uses the predicted base-layer tokens to progressively generate sequences for the remaining quantization layers. Finally, all tokens are transformed back into motion sequences via the RVQ-VAE decoder.

\paragraph{Training Objective.}
To train our gesture generation models, $\mathcal{L}_{mask}$, and  $\mathcal{L}_{res}$ functions for two generaors respectively by minimizing the categorical cross-entropy loss, as illustrated below:
\begin{equation}
\mathcal{L}_{mask} = \sum_{i=1}^{T} -\log p_\phi(t_i | Es(S), \mathrm{MASK}), \quad \mathcal{L}_{res} = \sum_{j=1}^{V}\sum_{i=1}^{T}-\log p_\phi(t_i^j | t_i^{1:j-1}, Es(S), j).
\end{equation}
In this formulation, $\mathcal{L}_{mask}$ predicts the masked motion tokens $t_i$ at each time step $i$ based on the input audio and the special $[\mathrm{MASK}]$ token. Conversely, $\mathcal{L}_{res}$ focuses on learning from multiple quantization layers, where $t_i^j$ represents the motion token from quantizer layer $j$ and $t_i^{1:j-1}$ includes the tokens from preceding layers. We also feed the predicted tokens into the RVQ decoder for gesture reconstructions, with velocity and acceleration losses~\cite{mdm, siyao2022bailando}.


\section{Speech-Gesture Retrieval}
\label{sec:s2g-alignment}

To validate the effectiveness of Speech-Gesture Alignment, inspired by TMR~\cite{tmr} we propose the Speech-Gesture Retrieval as the evaluation benchmark.

\noindent\textbf{Evaluation Settings.}
The retrieval performance is measures under recall at various ranks, R@1, R@2, etc.\. Recall at rank $k$ indicates the percentage of times the correct label is among the top $k$ results; therefore higher is better. We define two settings. The retrieval is based on the sliced clips, with each lasting for 4 seconds and 120 frames, in total 8176 samples.

(a) \textbf{All} test set samples for face and body motions. This set is problematic because the speech and gesture motion should not be of one-to-one mapping relationship.

(b) \textbf{Small batch} size of 32 speech-gesture pairs are randomly picked, reporting average performance.

\noindent\textbf{Evaluation Result.}
Shown in \cref{tab:s2g-alignment}, the gesture patterns and speech context do not present one-to-one mapping relationship, leading to the significantly low performance of retrieval. However, based on setting (b), within a small batch size of 32, the model achieves significantly higher performance, indicating the alignment provides the discrimination over different speech context and the motion. In addition, chronological negative examples during contrastive training enhances the robustness of retrieval.

\begin{table*}
    \centering
    \setlength{\tabcolsep}{4pt}
    \caption{\small{\textbf{Speech-to-Gesture Motion retrieval benchmark on PATS:}
    We establish two evaluation settings as described in
    Section~\ref{sec:s2g-alignment}. 
    }}
    \label{tab:s2g-alignment}
    \resizebox{0.99\linewidth}{!}{
    \begin{tabular}{l|ccccc|ccccc}
        \toprule
        
         \multirow{2}{*}{\textbf{Setting}}  & \multicolumn{5}{c|}{Speech-Face retrieval} & \multicolumn{5}{c}{Face-Speech retrieval} \\
         
         & \small{R@1 $\uparrow$} & \small{R@2 $\uparrow$} & \small{R@3 $\uparrow$} &  \small{R@5 $\uparrow$} & \small{R@10 $\uparrow$} & \small{R@1 $\uparrow$} & \small{R@2 $\uparrow$} & \small{R@3 $\uparrow$} & \small{R@5 $\uparrow$} & \small{R@10 $\uparrow$} \\
        \midrule
    
        (a) All  &  0.181 &  0.350 & 0.485 &  0.722 &  1.343 &  0.226  & 0.361   &0.429   &0.677   &1.207 \\
        \midrule

        (a) + chrono &  0.231  & 0.372  & 0.501  & 0.734  & 1.696 & 0.323 & 0.398  & 0.454 & 0.712 & 1.332 \\
        
        \midrule
        
        (b) Small batches &  26.230  & 45.318  & 59.330  & 77.019   &89.858 & 24.977 &44.822  &59.894 &77.775 &90.264 \\
        \midrule
        (b) + chrono & 27.437  & 47.552  & 63.193  & 74.343   & 89.996 & 26.451 & 46.432 & 61.727 & 79.779 & 91.373 \\

        \midrule

        \multirow{2}{*}{\textbf{Setting}}  & \multicolumn{5}{c|}{Speech-Body retrieval} & \multicolumn{5}{c}{Body-Speech retrieval} \\
         
         & \small{R@1 $\uparrow$} & \small{R@2 $\uparrow$} & \small{R@3 $\uparrow$} &  \small{R@5 $\uparrow$} & \small{R@10 $\uparrow$} & \small{R@1 $\uparrow$} & \small{R@2 $\uparrow$} & \small{R@3 $\uparrow$} & \small{R@5 $\uparrow$} & \small{R@10 $\uparrow$}\\
        \midrule
    
        (a) All  &  0.102 &  0.237 & 0.327 &  0.587 &  1.230 &  0.158  & 0.271   &0.406   &0.654   &1.320 \\
        \midrule

        (a) + chrono  &  0.132 &  0.257 & 0.373 &  0.603 &  1.340 &  0.178  & 0.289   &0.443   &0.671   &1.404 \\
        \midrule
    
        (b) Small batches &  25.542  & 43.660  & 57.954  & 77.471   &90.309 & 24.052 & 43.874 & 58.495 & 76.986 & 89.745 \\
        \midrule
        (b) + chrono & 28.732  & 48.569  & 59.958  &79.321 & 90.003 & 22.671 & 45.737 & 57.669 & 79.565 & 90.672 \\
                                      
    \bottomrule        
    \end{tabular}

    }
    
\end{table*}

\section{Additional Experiments}
\label{sec:sub-exp}

In the main paper, we have shown our method achieves promising joint gesture motion and video generation. To understand the disentangled gesture and video avatar generation separately, we further conduct Video Avatar Animation experiments separately to compare our method with the corresponding representative works.

\begin{figure}[t]
    \centering
    \includegraphics[width=0.9\linewidth]{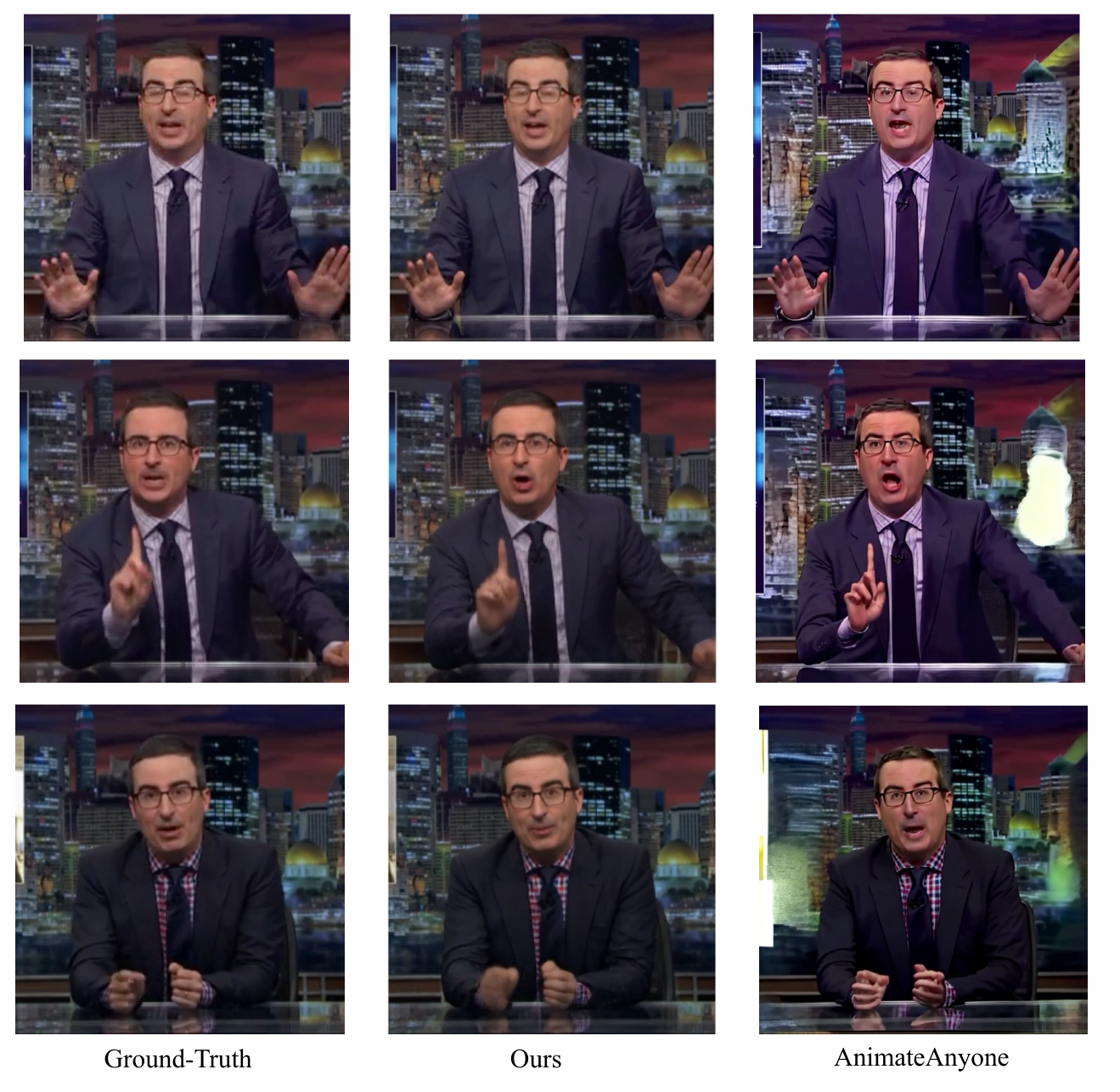}
    \caption{\small{\textbf{Comparison of Video Avatar Animation} Though presented with worse hand structure reconstruction, we achieve better identity preserving and significantly better background motion.}}
    \label{fig:image-reconst}
\end{figure}

\subsection{Video Avatar Animation}
\noindent\textbf{Experiment Settings.}
We select PATS dataset as in main paper for avatar rendering comparison. We processed the videos into 512x512 for Diffusion-bassed model AnimateAnyone~\cite{hu2023animateanyone}. We extract the 2D poses by MMPOse~\cite{mmpose2020} for pose guidance for the Diffusion Model, and maintain all the training details as in AnimateAnyone for consistency.

\noindent\textbf{Experiment Results.}
We compare the gesture generation module of our work with representative AnimateAnyone~\cite{hu2023animateanyone}. As shown in \cref{fig:image-reconst}, though AnimateAnyone achieves better video generation quality for hand structure of the speaker centering in the video, it fails to maintain the speaker identity, making the avatar less similar to the source image compared with our method. In addition, due to the entanglement of camera motions and speaker gesture motions within the dataset, AnimateAnyone fails to separate two types of motions from the source training video, thus leading to significant background changes over time and dynamic inconsistency. Unlike completely relying on human skeletons as conditions in AnimateAnyone, our method benefits from Warping-based method, which has the capability of resolving the background motions in addition to the speaker motion. We defer visual comparisons in the Appendix videos.

\section{Time and Resource Consumption}
\label{sec:sub-time}

In \cref{tab:consumption}, we present a comparison of training and inference times against existing baseline methods. For audio-gesture generation, our model's training time is comparable, albeit slightly slower, than that of ANGIE~\cite{angie} and S2G-Diffusion~\cite{s2gdiffusion}, primarily due to the inclusion of additional modules. However, it is considerably faster than MM-Diffusion~\cite{ruan2022mmdiffusion}. Notably, our method excels in inference speed, outperforming all other baselines.

While the training of image-warping and image refinement requires a lot of time, our method leads to a substantial reduction in overall time and resource usage compared to MM-Diffusion and other stable-diffusion-based video generation approaches. Furthermore, the generative masking paradigm we employ significantly cuts down inference times when compared to diffusion-based models like S2G-Diffusion or the autoregressive generations in ANGIE.

We further compared image-warping based method computation requirements with Stable Diffusion-based models like AnimateAnyone~\cite{hu2023animateanyone} in \cref{tab:consumption-sd}.

\begin{table}[!ht]
\centering
\caption{\small{\textbf{Time consumption comparison} of training (1 NVIDIA A100 GPU) and inference (1 NVIDIA GeForce RTX A6000 GPU).}}
\label{tab:consumption}
\resizebox{\columnwidth}{!}{%
\begin{tabular}{cccc}
\toprule
Name & Training & Training Breakdown & \begin{tabular}[c]{@{}c@{}} Inference\\ (video of $\sim$10 sec) \end{tabular} \\
\midrule
ANGIE & $\sim$5d & Motion Repr. $\sim$3d + Quantize $\sim$0.2d + Gesture GPT $\sim$1.8d  & $\sim$30 sec\\
MM-Diffusion \ & $\sim$14d  & Generation $\sim$9d + Super-Resolution $\sim$5d& $\sim$600 sec \\
S2G-Diffusion & $\sim$5d & Motion Decouple $\sim$3d + Motion Diffusion $\sim$1.5d + Refine $\sim$0.5d & $\sim$35 sec\\
Ours & $\sim$6d & Quantize $\sim$0.2d + Mask-Gen $\sim$1.5d + Res-Gen $\sim$0.5d + Img-warp \& Refine $\sim$3.5d  & $\sim$3 sec\\
\bottomrule
\end{tabular}%
}
\end{table}

\begin{table}[!ht]
\centering
\caption{\small{\textbf{Resource consumption comparison} with Stable-Diffusion-based Image-Animation models (1 NVIDIA A100 GPU), * means our re-implementation on PATS dataset.}}
\label{tab:consumption-sd}
\begin{center}
\resizebox{\columnwidth}{!}{
\begin{tabular}{ccccccc}
\toprule  
Methods & Training$\downarrow$ & Batch Size & Resolution & Memory$\downarrow$ & Training Task & Inference$\uparrow$  \\
\midrule  
AnimateAnyone* & 10 days & 4 & 512 & 44 GB & Pose-2-Img & -\\
AnimateAnyone* & 5 days & 4  & 512 & 36GB & Img-2-Vid & 15s \\
\midrule  
Ours & 2.5 days & 64 & 256 & 64 GB & Img-Warp & $\leq$1s\\
Ours & 1 day & 64 &  256 & 48GB & Img-Refine & $\leq$ 1s\\
\midrule  
Ours & 3.5 days & 32 & 512 &  60GB & Img-Warp & $\leq$1s\\
Ours & 1 day & 32 & 512 & 40GB & Img-Refine & $\leq$1s\\
\bottomrule  
\end{tabular}
}
\end{center}
\label{table_run_time}  
\end{table}

\begin{figure}
    \includegraphics[width=\linewidth]{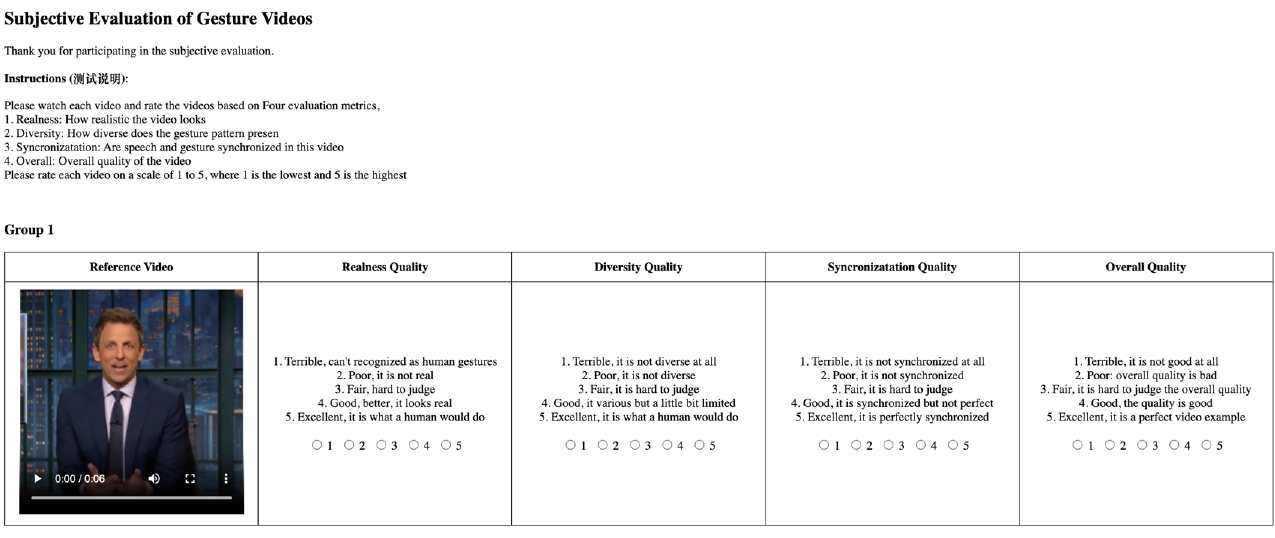}
    \caption{\small{Screenshot of user study website.}}
    \label{fig:user_study}
\end{figure}

\section{User Study Details}
\label{sec:sub_user}
For user study, we recruited 20 participants with good English proficiency. To conduct the user study, we randomly select 80 videos from EchoMimicV2~\cite{meng2024echomimic}, ANGIE~\cite{angie}, S2G-Diffusion~\cite{s2gdiffusion} and ours. Each user works on 20 videos, with 4 videos from each of the aforementioned methods. The users are not informed of the source of the video for fair evaluations. A visualization of the user study is shown in ~\cref{fig:user_study}.

\section{TPS-based Image-Warping}
\label{sec:sub-tps}

In this paper, we utilize Thin Plate Splines (TPS)~\cite{zhao2022thin} to model deformations based on human poses for image-warping. Here, we provide additional details on this approach.

The TPS transformation accepts \(N\) pairs of corresponding keypoints \((p^\mathbf{D}_i, p^\mathbf{S}_i)\) for \(i=1,2,\ldots, N\) (referred to as control points) from a driving image \(\mathbf{D}\) and a source image \(\mathbf{S}\). It outputs a pixel coordinate mapping \(\mathcal{T}_{tps}\left(\cdot\right)\), which represents the backward optical flow from \(\mathbf{D}\) to \(\mathbf{S}\). This transformation is founded on the principle that 2D warping can be effectively modeled through a thin plate deformation mechanism. The TPS transformation seeks to minimize the energy associated with bending this thin plate while ensuring that the deformation aligns accurately with the control points. The mathematical formulation is as follows:
\begin{equation}
\begin{split}
\min &\iint_{\mathbb{R}^{2}} \left( \left( \frac{\partial^{2} \mathcal{T}_{tps}}{\partial x^{2}} \right)^{2} + 2 \left( \frac{\partial^{2} \mathcal{T}_{tps}}{\partial x \partial y} \right)^{2} + \left( \frac{\partial^{2} \mathcal{T}_{tps}}{\partial y^{2}} \right)^{2} \right) \, dx dy, \label{eq:tps_1} \\
&\text{s.t.} \quad \mathcal{T}_{tps}(p_{i}^{\mathbf{D}}) = p_{i}^{\mathbf{S}}, \quad i = 1,2, \ldots, N,
\end{split}
\end{equation}
where \(p_{i}^{\mathbf{D}}\) and \(p_{i}^{\mathbf{S}}\) denote the \(i^{th}\) keypoints in \(\mathbf{D}\) and \(\mathbf{S}\) respectively. As shown in \cite{zhao2022thin}, it can be demonstrated that the TPS interpolating function satisfies \cref{eq:tps_1}:
\begin{equation}
\mathcal{T}_{tps}(p)=A\left[\begin{array}{l}
p \\
1
\end{array}\right]+\sum_{i=1}^{N} w_{i} U\left(\left\lVert p^{\mathbf{D}}_{i}-p\right\rVert_{2}\right), \label{eq:tps_2}
\end{equation}
where \(p=(x, y)^{\top}\) represents the coordinates in \(\mathbf{D}\), and \(p^{\mathbf{D}}_{i}\) is the \(i^{th}\) keypoint in \(\mathbf{D}\). The function \(U(r)=r^{2} \log r^{2}\) serves as a radial basis function. Notably, \(U(r)\) is the fundamental solution to the biharmonic equation \cite{selvadurai2000biharmonic}, defined by:
\begin{equation}
\Delta^2 U=\left(\frac{\partial^2}{\partial x^2}+\frac{\partial^2}{\partial y^2}\right)^2 U \propto \delta_{(0,0)},
\end{equation}
where the generalized function \(\delta_{(0,0)}\) is characterized as:
\begin{align}
\delta_{(0,0)} = \begin{cases}
\infty, & \text{if } (x, y) = (0, 0) \\
0, & \text{otherwise}
\end{cases},\quad
\text{and} \iint_{\mathbb{R}^{2}} \delta_{(0,0)}(x, y) \,dx dy = 1,
\end{align}
indicating that \(\delta_{(0,0)}\) is zero everywhere except at the origin, where it integrates to one.

We denote the \(i^{th}\) keypoint in image \(\mathbf{X}\) (either \(\mathbf{D}\) or \(\mathbf{S}\)) as \(p^{\mathbf{X}}_i=(x^{\mathbf{X}}_i, y^{\mathbf{X}}_i)^{\top}\), and we define:
\begin{equation}
r_{ij} = \left\lVert p^{\mathbf{D}}_{i}-
p^{\mathbf{D}}_{j}\right.\rVert, \quad i,j=1,2,\ldots, N. 
\end{equation}
Next, we construct the following matrices:
\begin{equation}
\begin{aligned} 
K= & {\left[\begin{array}{cccc}0 & U\left(r_{12}\right) & \cdots & U\left(r_{1N}\right) \\
U\left(r_{21}\right) & 0 & \cdots & U\left(r_{2 N}\right) \\ \vdots & \vdots & \ddots & \vdots \\
U\left(r_{N 1}\right) & U\left(r_{N 2}\right) & \cdots & 0\end{array}\right] }, \quad
P =\left[\begin{array}{ccc}1 & x^{\mathbf{D}}_1 & y^{\mathbf{D}}_1 \\ 1 & x^{\mathbf{D}}_2 & y^{\mathbf{D}}_2 \\ \vdots & \vdots & \vdots \\ 1 & x^{\mathbf{D}}_N & y^{\mathbf{D}}_N\end{array}\right],
\end{aligned}
\end{equation}
\begin{equation}
\begin{aligned}
L & =\left[\begin{array}{cc}K & P \\ P^T & 0\end{array}\right], \quad
Y = \left[\begin{array}{ccccccc} x^{\mathbf{S}}_1 & x^{\mathbf{S}}_2 & \cdots &  x^{\mathbf{S}}_N & 0 & 0 & 0 \\
y^{\mathbf{S}}_1 & y^{\mathbf{S}}_2 & \cdots &  y^{\mathbf{S}}_N & 0 & 0 & 0 \\
\end{array}\right]^{\top}. 
\end{aligned}
\end{equation}
We can then determine the affine parameters \(A \in \mathcal{R}^{2 \times 3}\) and the TPS weights \(w_{i} \in \mathcal{R}^{2 \times 1}\) by solving the following equation:
\begin{equation}
\begin{aligned}
\left[w_1, w_2, \cdots, w_N, A\right]^{\top} = L^{-1}Y.
\end{aligned}
\end{equation}
In \cref{eq:tps_2}, 
the first term \(A\left[\begin{array}{l} 
p \\ 1 \end{array}\right]\) represents 
an affine transformation that aligns the paired control points \((p^\mathbf{D}_i, p^\mathbf{S}_i)\) in linear space.
The second term \(\sum_{i=1}^{N} w_{i} U\left(\left\lVert p^{\mathbf{D}}_{i}-p\right\rVert_{2}\right)\) accounts for nonlinear distortions that enable the thin plate to be elevated or depressed. 
By combining both linear and nonlinear transformations, the TPS framework facilitates precise deformations, which are essential for accurately capturing motion while preserving critical appearance details within our framework.

\section{Ethical Considerations} 
\label{sec:ethics}

While this work is centered on generating co-speech gesture videos, it also raises important ethical concerns due to its potential for photo-realistic rendering. This capability could be misused to fabricate videos of public figures making statements or attending events that never took place. Such risks are part of a broader issue within the realm of AI-generated photo-realistic humans, where phenomena like deepfakes and animated representations pose significant ethical challenges.

Although it is difficult to eliminate the potential for misuse entirely, our research offers a valuable technical analysis of gesture video synthesis. This contribution is intended to enhance understanding of the technology's capabilities and limitations, particularly concerning details such as facial nuances and temporal coherence.

In addition, we emphasize the importance of responsible use. We recommend implementing practices such as watermarking generated videos and utilizing synthetic avatar detection tools for photo-realistic images. These measures are vital in mitigating the risks associated with the misuse of this technology and ensuring ethical standards are upheld.

\section{Limitations}
\label{sec:limitation}

While our method have achieved significant improvements over existing baselines, there are still two limitations of the current work.

First, the generation quality still exhibit blurries and flickering issues. The intricate structure of hand hinders the generator in understanding the complex motions. In addition, PATS dataset is sourced from in-the-wild videos of low quality. Most frames extracted from videos demonstrate blurry hands, limiting the network learning. Thus, it is important to collect the high-quality gesture video dataset with clearer hands to further enhance the generation quality.

Second, when modeling the whole upper-body, it is hard to achieve synchronized lip movements aligned with the audio. Even though we explicit separate the face motion and body motion to deal with this problem, there is no regularization on lip movement. We would like to defer this problem to the future works that models disentangled and fine-grained motions for each face and body region.

\end{document}